\documentclass[10pt,twocolumn,letterpaper]{article}

\usepackage{cvpr}
\usepackage{times}
\usepackage{epsfig}
\usepackage{graphicx}
\usepackage{amsmath}
\usepackage{amssymb}

\usepackage{url}
\usepackage{xcolor}
\usepackage{graphicx}
\usepackage{amsmath}
\usepackage{amssymb}
\usepackage{url}
\usepackage{xcolor}
\definecolor{newcolor}{rgb}{.8,.349,.1}
\usepackage[fleqn]{mathtools}
\usepackage{enumerate}
\usepackage{enumitem}
\usepackage[export]{adjustbox}
\usepackage{framed}
\usepackage{subcaption}
\usepackage[export]{adjustbox}
\usepackage{xcolor}
\usepackage{array}
\usepackage{pifont}
\usepackage{soul}
\usepackage{xspace}
\usepackage{float}
\usepackage{placeins}
\definecolor{newcolor}{rgb}{.8,.349,.1}

\usepackage{placeins}

\usepackage[breaklinks=true,bookmarks=false]{hyperref}

\cvprfinalcopy 

\def\cvprPaperID{****} 

\begin{document}

\title{DeepObjStyle: Deep Object-based Photo Style Transfer}

\author{Indra Deep Mastan and Shanmuganathan Raman\\
Indian Institute of Technology Gandhinagar\\
Gandhinagar, Gujarat, India\\
{\tt\small \{indra.mastan, shanmuga\}@iitgn.ac.in}}
\maketitle

\begin{abstract}
One of the major challenges of style transfer is the appropriate image features supervision between the output image and the input (style and content) images. An efficient strategy would be to define an object map between the objects of the style and the content images.  However, such a mapping is not well established when there are semantic objects of different types and numbers in the style and the content images. It also leads to content mismatch in the style transfer output, which could reduce the visual quality of the results. We propose an object-based style transfer approach, called DeepObjStyle, for the style supervision in the training data-independent framework. DeepObjStyle preserves the semantics of the objects and achieves better style transfer in the challenging scenario when the style and the content images have a mismatch of image features. We also perform style transfer of images containing a word cloud to demonstrate that DeepObjStyle enables an appropriate image features supervision. We validate the results using quantitative comparisons and user studies.
\end{abstract}

\begin{figure*}[!h]
\begin{center}
\begin{subfigure}{0.142\textwidth} \includegraphics[width=\linewidth]{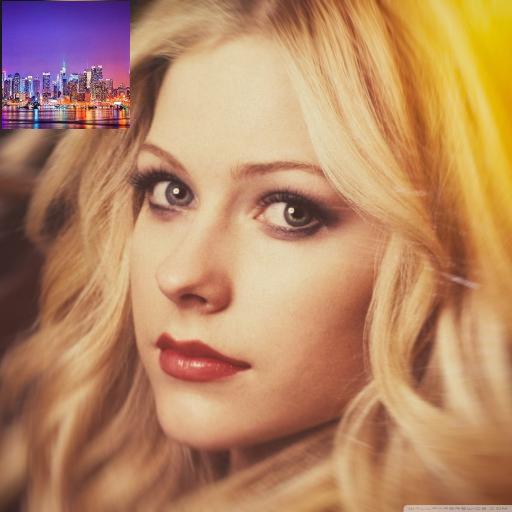} \end{subfigure}
\begin{subfigure}{0.142\textwidth} \includegraphics[width=\linewidth]{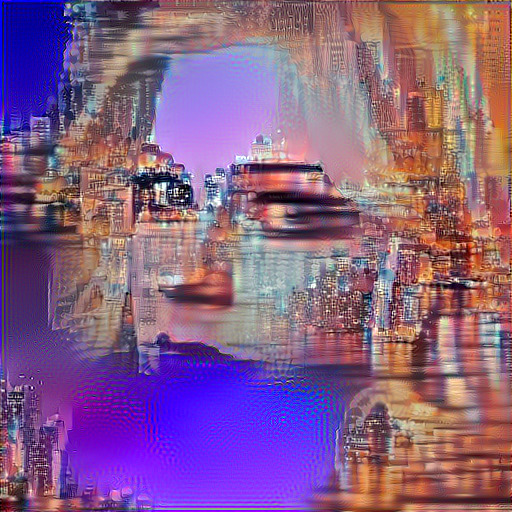}\end{subfigure}
\begin{subfigure}{0.142\textwidth} \includegraphics[width=\linewidth]{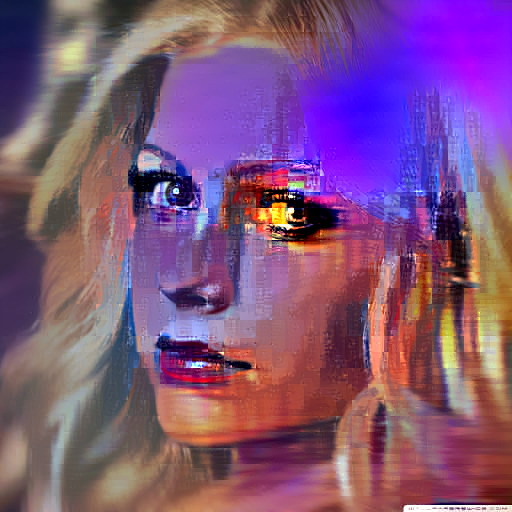} \end{subfigure} 
\begin{subfigure}{0.142\textwidth} \includegraphics[width=\linewidth]{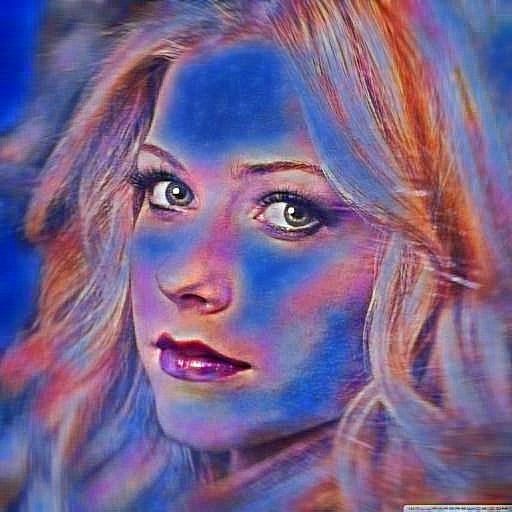} \end{subfigure}
\begin{subfigure}{0.142\textwidth} \includegraphics[width=\linewidth]{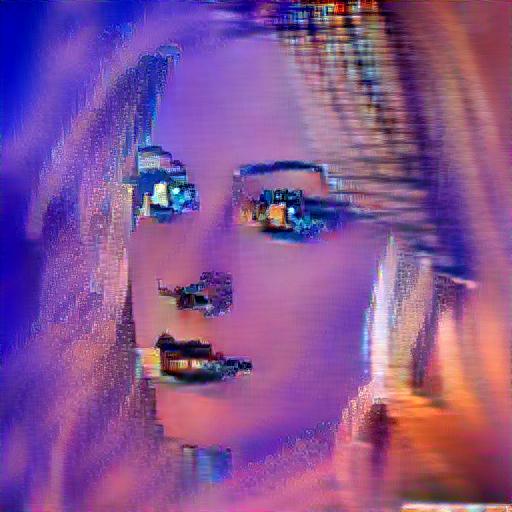} \end{subfigure}
\begin{subfigure}{0.142\textwidth} \includegraphics[width=\linewidth]{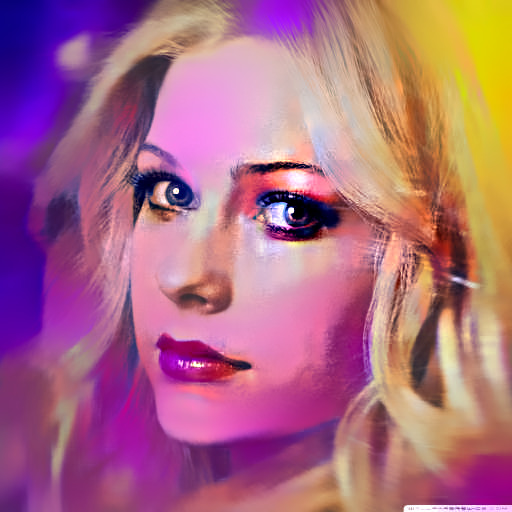}\end{subfigure} \\
\begin{subfigure}{0.142\textwidth} \includegraphics[width=\linewidth]{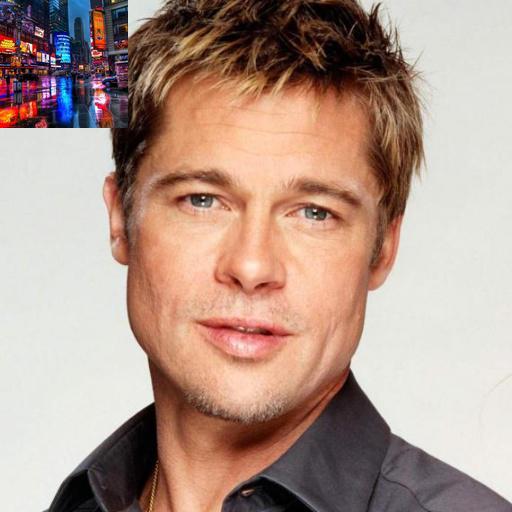} \caption{\small  Content \& Style } \end{subfigure}
\begin{subfigure}{0.142\textwidth} \includegraphics[width=\linewidth]{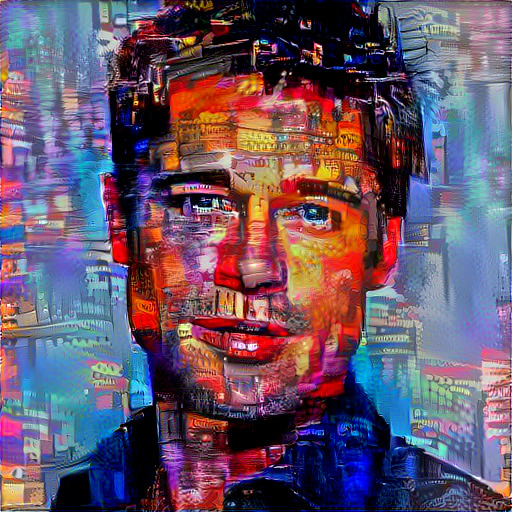}   \caption{\small  Neural Style \cite{gatys2016image} } \end{subfigure}
\begin{subfigure}{0.142\textwidth} \includegraphics[width=\linewidth]{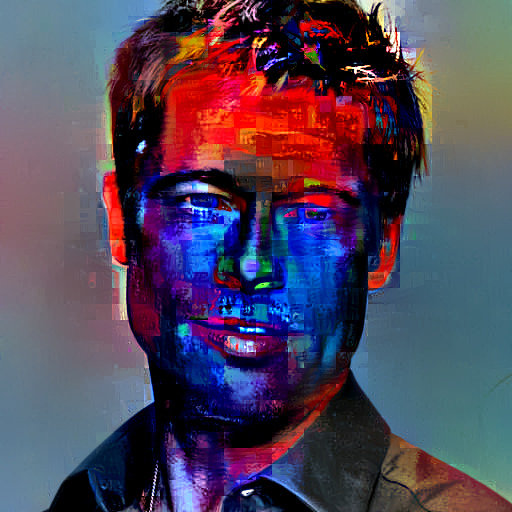}  \caption{\small  DPS \cite{luan2017deep}} \end{subfigure}
\begin{subfigure}{0.142\textwidth} \includegraphics[width=\linewidth]{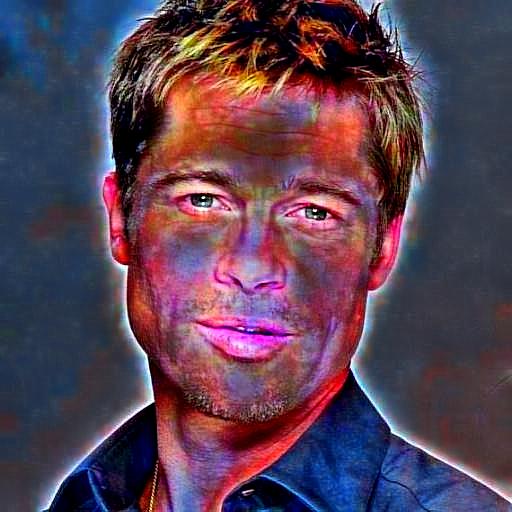}  \caption{\small  WCT2 \cite{yoo2019photorealistic}}  \end{subfigure}
\begin{subfigure}{0.142\textwidth} \includegraphics[width=\linewidth]{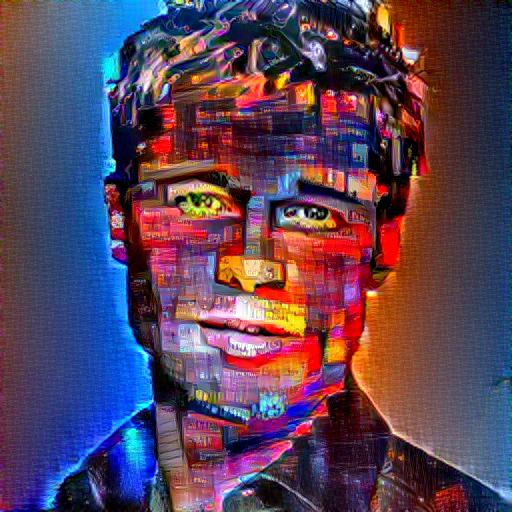}  \caption{\small  STROTSS \cite{kolkin2019style}} \end{subfigure}
\begin{subfigure}{0.142\textwidth} \includegraphics[width=\linewidth]{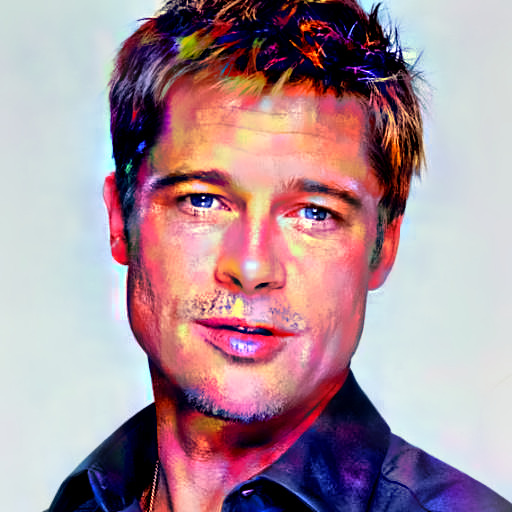}  \caption{\small  DeepObjStyle} 
 \end{subfigure}\vspace*{-0.15cm}
    \caption{The figure shows style transfer in the challenging scenario of the mismatch of image features between style and content images. The content image is the face of a person and the style image is the scene of the buildings (top left corner).  It could be observed that DeepObjStyle output images with better quality.}\label{fig: photo}
\end{center}%
\end{figure*}

\section{Introduction and Prior Work}
\label{sec: introduction}
Style transfer is an ill-posed problem that synthesizes a new image using the style and the content images \cite{gatys2015neural}. Style transfer output captures objects from the content image. The aim is to perform the image feature enhancement \cite{li2018lightennet, wang2020ebit, wang2016color, xie2020semantically, yin2020novel} of the content image. The objects of the style transfer output get the style features from the style image. Style transfer when the content and the style images have the same set of semantic objects enables applications such as puppet control and domain translation \cite{mechrez2018contextual}. Style transfer methods when the content and the style images have different sets of semantic objects are called arbitrary style transfer methods \cite{gu2018arbitrary}.

We classify deep convolutional neural network (CNN) based style transfer methods for simplicity as follows: (1) training data-based and (2) training data-independent methods. Training data-based style transfer methods use samples of the style and the content images to train the CNN. The training process would learn the style features from the style image and the content features from the content image \cite{chen2017stylebank}. Training-independent style transfer methods do not train a CNN with the samples of the style and the content images and mostly focus on designing the loss function for the deep features extracted using VGG19 \cite{gatys2016image, luan2017deep, kolkin2019style, gu2018arbitrary}.

Training-data based methods that use adversarial loss require many samples of the style and the content images  \cite{isola2017image}. Zhang \textit{et al.} have shown how to minimize the number of sample images by mixing the encoded latent representations of the style and the content images  \cite{zhang2018separating}. Mechrez \textit{et al.} proposed contextual loss (CL), which trains CNN using the samples of the content images \cite{mechrez2018contextual}. The stylization by Li \textit{et al.} \cite{li2018closed} uses Microsoft COCO dataset \cite{lin2014microsoft} to train the decoder network. WCT2 \cite{yoo2019photorealistic} uses images from Microsoft COCO dataset to improvise \cite{li2018closed}. 

Training data-independent setup includes the style transfer of arbitrary images using the \textit{gram} loss \cite{gatys2016image} proposed by Gatys \textit{et al.}. Luan \textit{et al.} improvised \cite{gatys2016image} for photo-realistic style transfer and geometric structure preservation  \cite{luan2017deep}. They validated the quality of style transfer output by checking style spillover between dissimilar objects, also known as the \emph{content mismatch} criterion. However, the content mismatch is inherent when the semantic objects in the style and the content images are of different types and numbers. Kolkin \textit{et al.} proposed a new technique using  optimal transport called STROTSS \cite{kolkin2019style}.


Image features transfer between contextually similar objects is a challenging task in training data-based and training data-independent setup \cite{li2017universal}. The training data-based methods have shown a good qualitative performance, but they could be biased towards the images in training samples and might not generalize well to the new style and content images. The training data-independent setting is challenging because of the lack of object context information learned from the sample images.

\begin{figure*}[!h] \begin{center}
    \begin{subfigure}[b]{0.23\linewidth}
        \includegraphics[width=\textwidth]{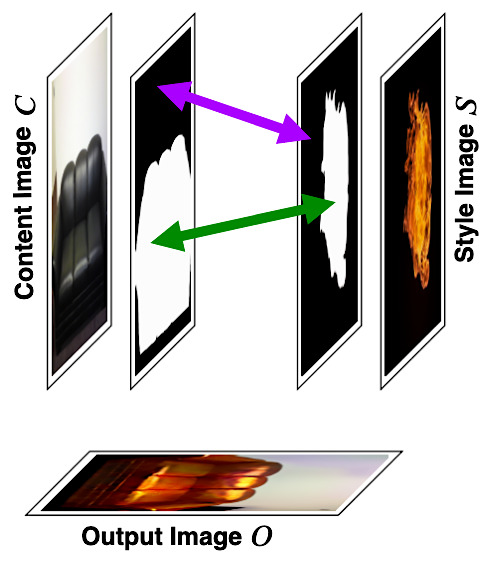} \caption{\small   \textbf{STP-E}, $``m=n"$.\\ $(m=2, n=2)$}
    \end{subfigure} \hspace*{0.75cm}
    \begin{subfigure}[b]{0.23\linewidth}
            \includegraphics[width=\textwidth]{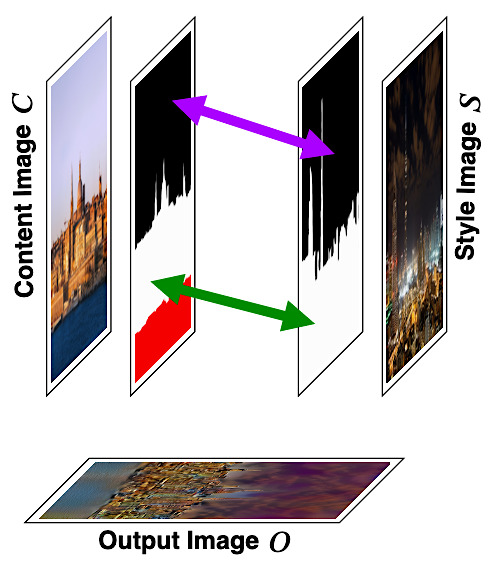} \caption{\small  \textbf{STP-C}, $``m>n"$.\\ $(m=3, n=2)$}
    \end{subfigure}  \hspace*{0.75cm}
    \begin{subfigure}[b]{0.23\linewidth}
\includegraphics[width=\textwidth]{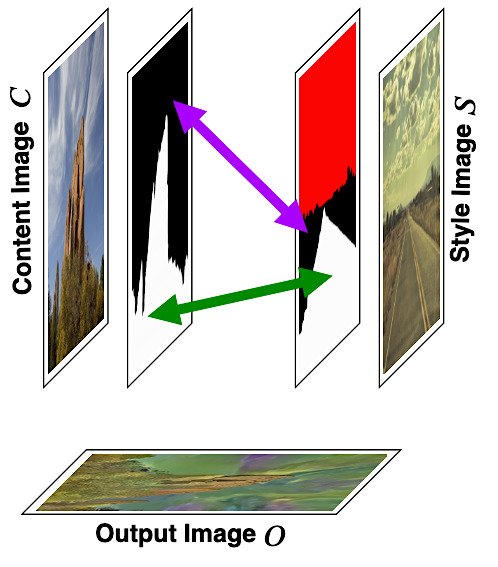} \caption{\small  \textbf{STP-S}, $``m<n"$.\\ $(m=2, n=3)$}
    \end{subfigure}%
     \end{center}\vspace*{-0.35cm}
    \caption{\textbf{Style Transfer Problems (STP).} The figure illustrates STP. The style image $S$ has $n$ semantic objects, and the content image $C$ has $m$ semantic objects. The arrows show the object mapping and the red color in the segmentation mask shows the unmapped object. (a) STP-E: style transfer when the objects in $S$ and  $C$ are equal. (b) STP-C: style transfer when $C$ has more objects than $S$. (c) STP-S: style transfer when $S$ has more objects than $C$. We describe STP in Sec.~\ref{sec: background}. }\label{fig: mappingPic}
\end{figure*}

Another challenge in style transfer is when the style and the content images have different types or different numbers of objects (Fig.~\ref{fig: mappingPic}).  One strategy would be to merge objects with similar classes (\textit{e.g.,} grass and tree) to minimize the content mismatch \cite{luan2017deep}. However, the strategy above would not be useful when the semantic objects are of different classes  (Sec.~\ref{sec: background}).

We address the style transfer challenges by an object-based style transfer approach and investigate how to utilize features from all the objects in the style and the content image while minimizing the content mismatch. We call our deep object-based photo-style transfer method as~DeepObjStyle. The strategy is to distribute the image features considering the semantics of the image features in the style and the content images. The object-based approach aims to enable image features supervision based on the contextually similar regions of the output.

DeepObjStyle achieves good perceptual quality in the style transfer output in the presence of different numbers of objects present in the style and the content images and when there is a mismatch of image features between the style and the content images. We also investigate the approach by putting  word cloud in the style and the content images. The word cloud would not be readable when the unrelated image features are merged with it. Therefore, it helps to investigate the preservation of the structure of the objects in the output. 

DeepObjStyle supervises features by minimizing distortions and considering the object context of the objects of the style and the content images. We propose unmapped object loss and mapped object loss for the distribution of image features to the style transfer output. 

\textbf{Contributions.} The major contributions of the work are mentioned below. 
\begin{enumerate}[noitemsep, leftmargin=*] %
\item DeepObjStyle preserves the semantics of the content image in the style transfer output (Fig.~\ref{fig: photo}). 
\item We investigate the performance of DeepObjStyle in the following two challenging scenarios. Style transfer when there is a low similarity between the features of the style and the content images (Fig.~\ref{fig: texture} and Fig.~\ref{fig: mismatch}). Style transfer where the style and the content images contain a word-cloud (Fig.~\ref{fig: wordcloud}). 
\item DeepObjStyle provides feature transfer from all the objects of the style and the content images while preserving the object structure (Fig.~\ref{fig: mappingPic}, Fig.~\ref{fig: wordcloud}, Fig.~\ref{fig: diffusion}, and Fig.~\ref{fig: utilization}).
\item We evaluated our results using reference-based quality assessment: Pieapp \cite{prashnani2018pieapp}, and no-reference quality assessment: NIMA \cite{idealods2018imagequalityassessment}. We also performed a user study to compare the style transfer outputs. We show that DeepObjStyle outperforms the relevant state-of-the-art methods (Sec.~\ref{sec: experiments}).
\end{enumerate}

\section{Background} \label{sec: background} 
\noindent \textbf{Content Mismatch.} We describe the content mismatch challenge of style transfer using an example as follows. Given a content image $C$ and a reference style image $S$, the objective is to synthesize a new image $O$, which contains the content features from $C$ and the style features from $S$. Also, suppose that the style image $S$ has two semantic objects and the content image $C$ has three semantic objects (Fig.~\ref{fig: mappingPic}-a).  The style transfer output $O$ gets the content features (\textit{i.e.}, three objects) from $C$. The style features in these three objects in $O$ are received from the two objects in $S$. The interesting thing to note is that providing style features from two objects to the three objects in $O$ would introduce content mismatch as there would be one object in $S$, which transfer style features to multiple objects in $O$.

\noindent \textbf{Style Transfer Problems.} The task is to synthesize a new image $O$ by using style features from the style image $S$ and content features from the content image $C$. Suppose the content image $C$ has $m$ objects and the style image has $n$ objects. We have shown Style Transfer Problems (STP) in Fig.~\ref{fig: mappingPic}. We discuss STP in detail as follows. 
\begin{itemize}[noitemsep,wide=0pt]
\item \underline{STP-E}: Style transfer when the objects in content image and the style image are equal (\textit{i.e.}, $m=n$ in Fig.~\ref{fig: mappingPic}-a). Therefore, there exists a one-to-one object map between the objects of the style image and the objects of the content image. Moreover, STP-E is challenging when $C$ and $S$ have semantically different objects. It is due to the fact that the style transfer method has to synthesize a new image by mixing content features and style features of the different class of objects (\textit{i.e.}, content mismatch in Fig.~\ref{fig: mismatch}). We use a mapped object loss to transfer image features based on the object context for the above STP-E problem.
\item \underline{STP-C}: Style transfer when the content image has more semantic objects than that of the style image (\textit{i.e.}, $m>n$ in Fig.~\ref{fig: mappingPic}-b).  Therefore, a content mismatch occurs by the utilization of style features from multiple objects of $S$ to an object in $O$. We propose unmapped objects loss to minimize the content mismatch for the above style \emph{diffusion} problem (Fig.~\ref{fig: diffusion}).
\item \underline{STP-S}: Style transfer when the style image has more objects than that of the content image (\textit{i.e.}, $n>m$ in Fig.~\ref{fig: mappingPic}-c). Therefore, a content mismatch occurs due to the diffusion of style features from an object of $S$ to multiple objects in $O$. We propose unmapped objects loss to minimize the effects of a content mismatch for the above style \emph{utilization} problem (Fig.~\ref{fig: utilization}).
\end{itemize}

The style and the content features from $C$ and $S$ are extracted using the pre-trained VGG19 network \cite{simonyan2014very} denoted by $\phi$. We do not train CNN to learn image features from samples of the style and the content images. The extracted feature maps from $\phi$ are used to synthesize image features of the output $O$.


\section{Our Approach} \label{sec: dos}
\begin{figure*}[!h] \centering
\captionsetup[subfigure]{justification=centering}
    \begin{subfigure}[b]{0.17\linewidth} 
    \caption{Content \& Style}
        \includegraphics[width=\textwidth]{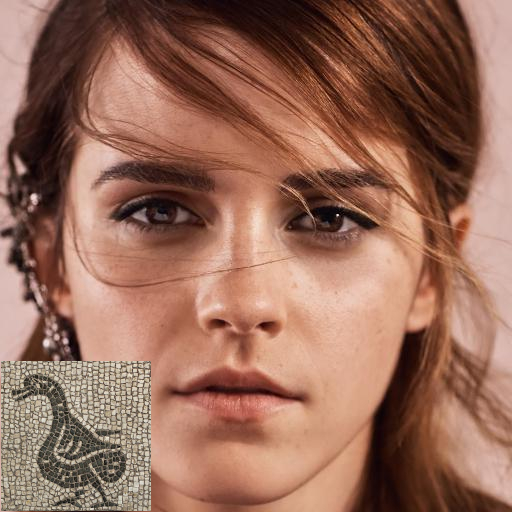} \vspace*{0.23cm}
    \end{subfigure}  \hspace*{1pt}%
    \begin{subfigure}[b]{0.17\linewidth} 
    \caption{DPS  \cite{luan2017deep}} 
        \includegraphics[width=\textwidth]{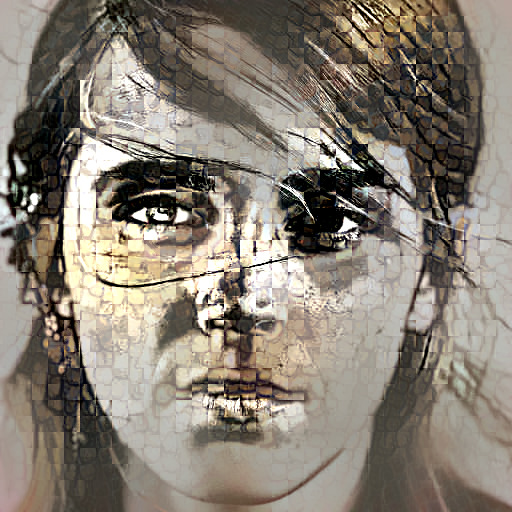} \caption*{ $E$: 5.86, $Q$:  5.40}
    \end{subfigure}  \hspace*{1pt}%
    \begin{subfigure}[b]{0.17\linewidth}
    \caption{WCT2 \cite{yoo2019photorealistic} }
        \includegraphics[width=\textwidth]{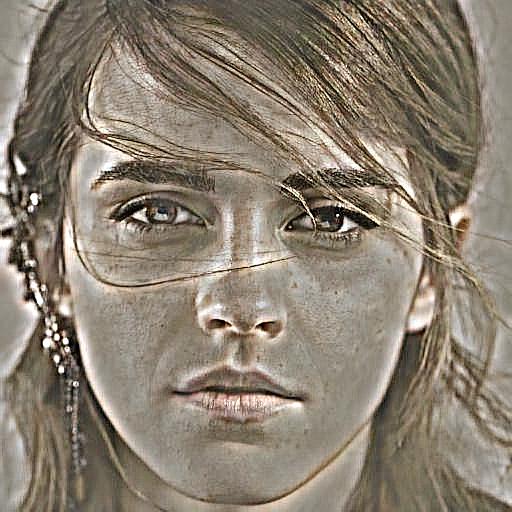} \caption*{ $E$: 5.45, $Q$:  5.20}
    \end{subfigure}  \hspace*{1pt}%
    \begin{subfigure}[b]{0.17\linewidth}
    	\caption{STROTSS  \cite{kolkin2019style}}
        \includegraphics[width=\textwidth]{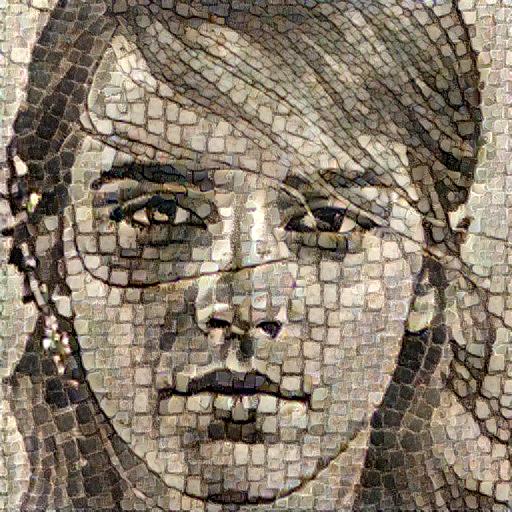} \caption*{ $E$: 6.52, $Q$:  5.22}
    \end{subfigure}  \hspace*{1pt}%
    \begin{subfigure}[b]{0.17\linewidth}
    \caption{DeepObjStyle}
        \includegraphics[width=\textwidth]{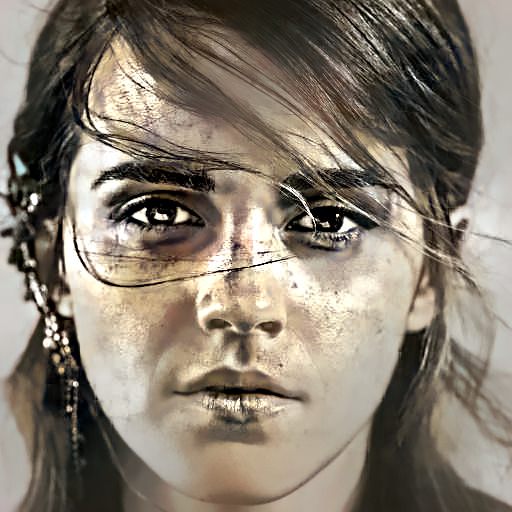} \caption*{ $E$: 4.65, $Q$:  5.83}
    \end{subfigure} \\
    \begin{subfigure}[b]{0.15\linewidth}
    \caption{\small  Content \& Style}    
        \includegraphics[width=\textwidth]{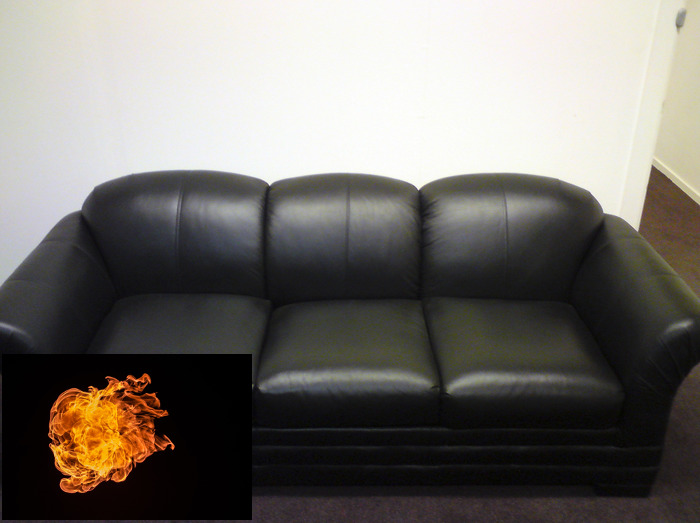} \vspace*{0.23cm}
    \end{subfigure}  \hspace*{1pt}%
    \begin{subfigure}[b]{0.15\linewidth}
    \caption{\small  DPS  \cite{luan2017deep}}     
        \includegraphics[width=\textwidth]{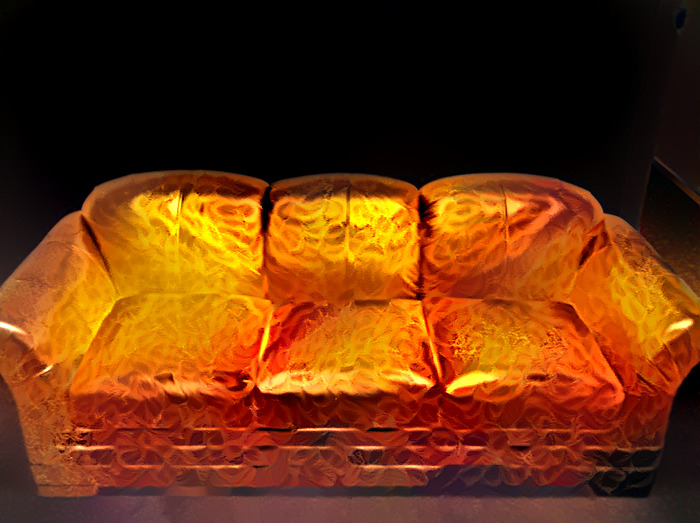} \caption*{ $E$: 4.03, $Q$:  4.78}
    \end{subfigure}  \hspace*{1pt}%
    \begin{subfigure}[b]{0.15\linewidth}
    \caption{\small  WCT2 \cite{yoo2019photorealistic} }    
        \includegraphics[width=\textwidth]{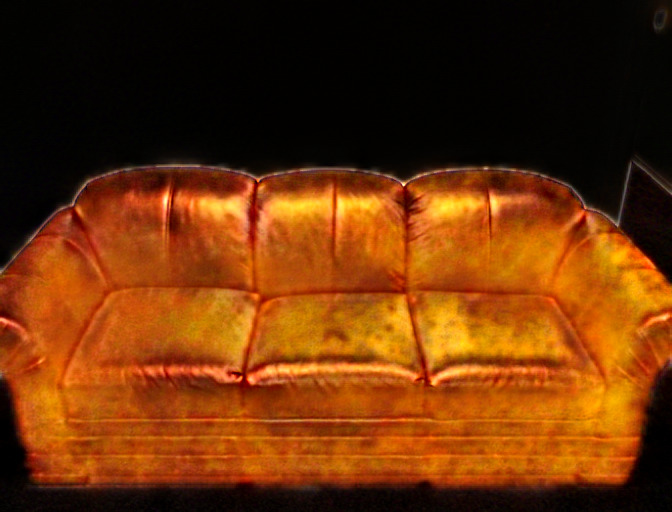} \caption*{ $E$: 3.77, $Q$:  4.45}
    \end{subfigure}  \hspace*{1pt}%
    \begin{subfigure}[b]{0.15\linewidth}
    	\caption{\small  STROTSS  \cite{kolkin2019style}}    
        \includegraphics[width=\textwidth]{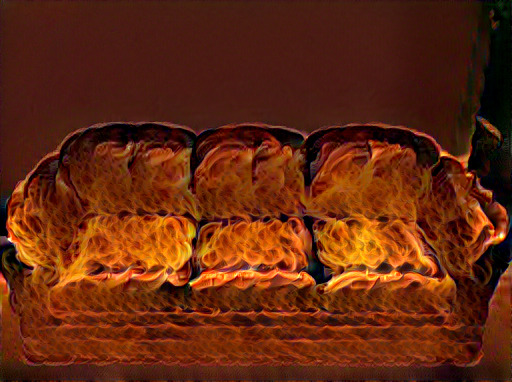} \caption*{ $E$: 4.18, $Q$:  4.83}
    \end{subfigure}  \hspace*{1pt}%
    \begin{subfigure}[b]{0.15\linewidth}
    \caption{\small  DeepObjStyle}    
        \includegraphics[width=\textwidth]{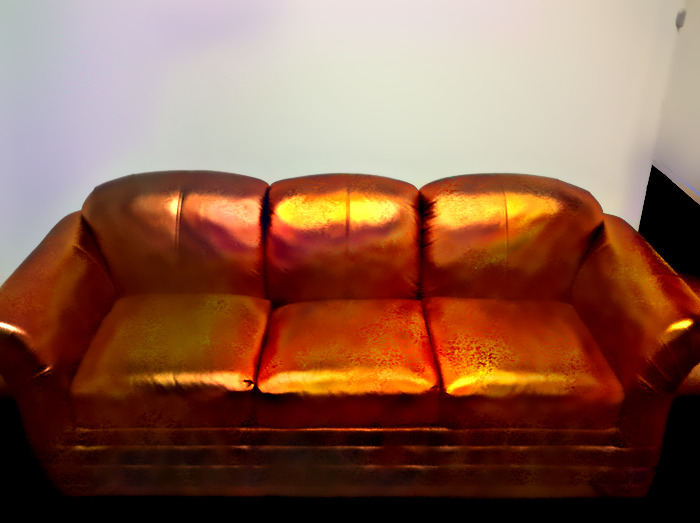} \caption*{ $E$: 2.7, $Q$: 5.16}
    \end{subfigure}  \vspace*{-0.15cm}
\caption{\textbf{Content Mismatch.} The style image is shown at the bottom left corner of the content image. These images have a content mismatch and challenging for style transfer \cite{luan2017deep}. The perceptual error score Pieapp \cite{prashnani2018pieapp} is denoted by $E$. The image quality score predicted by NIMA \cite{idealods2018imagequalityassessment} is denoted by $Q$. DeepObjStyle achieves a minimum perceptual error score $E$ and gets the highest quality score $Q$. DPS \cite{luan2017deep} and WCT2 \cite{yoo2019photorealistic} do not preserve the content features well. STROTSS  \cite{kolkin2019style} also suffers from the content mismatch. DeepObjStyle (ours) minimize the content mismatch and preserves the semantics of the objects.}\label{fig: mismatch}    
\end{figure*}

We have described that a good style transfer approach would allow features transfer while minimizing the effects of content mismatch and preserves the semantics of the objects in the style transfer output. One could perform style transfer without considering the objects, but it could result in low perceptual quality due to the mixing of features from different categories of objects in the output (Fig.~\ref{fig: mismatch}).

\textbf{Overview.} We first define an one-to-one object map (OM) between the contextually similar objects of the style and the content images using the segmentation mask. If the number of objects in the style and the content images are different, OM creates two categories of objects -  mapped objects and unmapped objects. Mapped objects are contained in both the style and the content images and participate in OM. Unmapped objects are contained either in the style image or in the content image and do not belong to OM. Next, we use the mapped objects loss and the unmapped objects loss to distribute features from the input style image and content image to the style transfer output. 

\textbf{DeepObjStyle loss.} We define DeepObjStyle loss $\mathcal{L}_{dos}$ as a combination of mapped objects loss $\mathcal{L}^{M}_{dos}$ and unmapped objects loss $\mathcal{L}^{U}_{dos}$ in  Eq.~\ref{eq: total}.
\begin{equation}\label{eq: total}
\mathcal{L}_{dos} = \alpha \mathcal{L}^{M}_{dos} + \beta \mathcal{L}^{U}_{dos} 
\end{equation}
Here, $\alpha$ and $\beta$ are the coefficients. The mapped objects loss $\mathcal{L}^{M}_{dos}$ uses  segmentation masks to define an one-to-one object map between the objects of the style and the content images for style transfer. The unmapped objects loss $\mathcal{L}^{U}_{dos}$ performs style transfer for unmapped objects independent of any mapping.

\begin{figure*}[!h]\centering
\begin{subfigure}[b]{0.148\textwidth} \includegraphics[width=\linewidth]{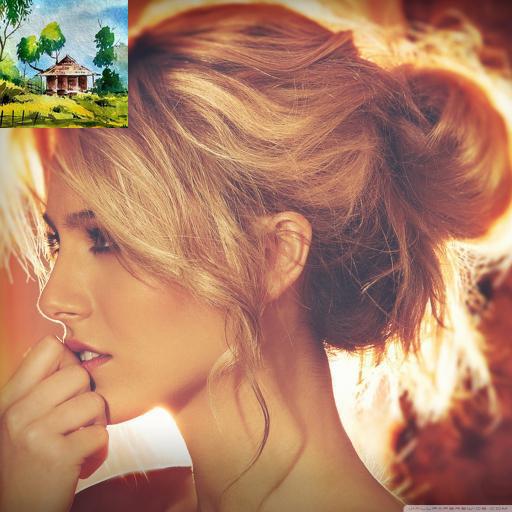} \caption{\small  Content \& Style}  \end{subfigure}
\begin{subfigure}[b]{0.148\textwidth} \includegraphics[width=\linewidth]{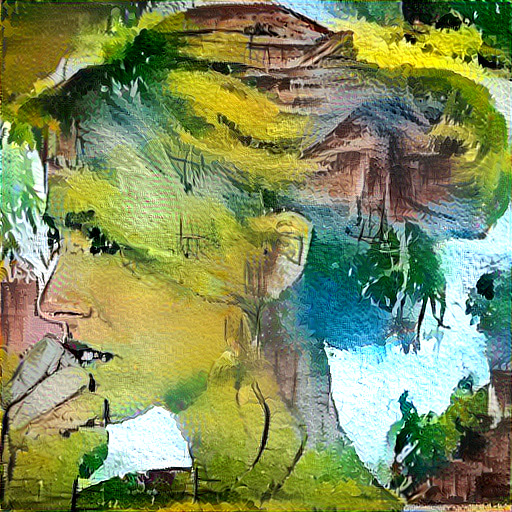} \caption{\small  Neural Style \cite{gatys2016image}} \end{subfigure}
\begin{subfigure}[b]{0.148\textwidth} \includegraphics[width=\linewidth]{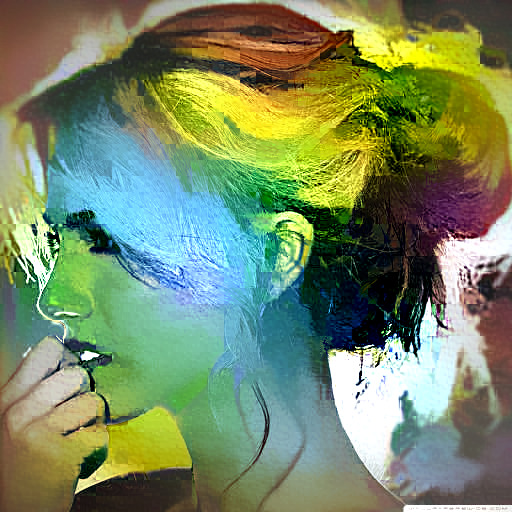} \caption{\small  DPS \cite{luan2017deep} } \end{subfigure}
\begin{subfigure}[b]{0.148\textwidth} \includegraphics[width=\linewidth]{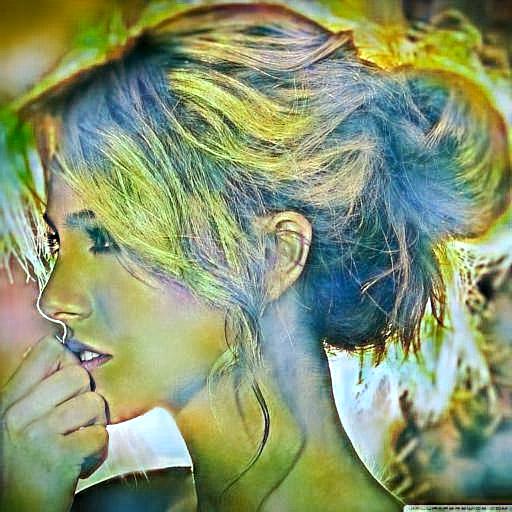} \caption{\small  WCT2 \cite{yoo2019photorealistic} } \end{subfigure}
\begin{subfigure}[b]{0.148\textwidth} \includegraphics[width=\linewidth]{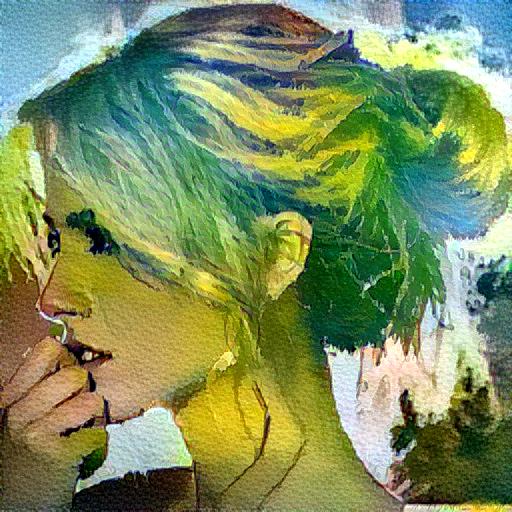} \caption{\small  STROTSS \cite{kolkin2019style}} \end{subfigure}
\begin{subfigure}[b]{0.148\textwidth} \includegraphics[width=\linewidth]{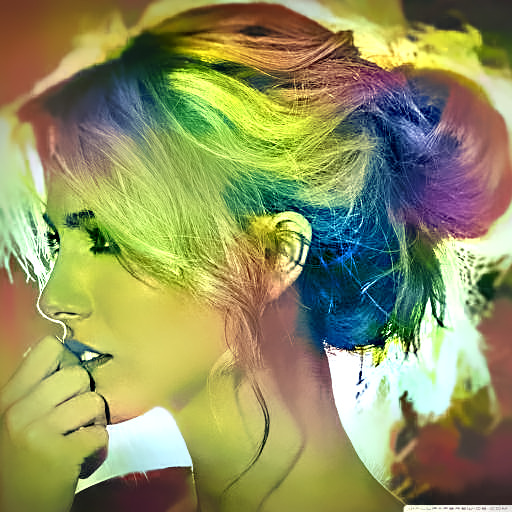} \caption{\small  DeepObjStyle} \end{subfigure} \vspace*{-0.15cm}
\caption{This figure shows the style transfer in the presence of the mismatch of image features where the style image is not photo-realistic, but the content image is photo-realistic. It could be observed that DeepObjStyle suppress distortions and preserve the photo-realism of the content image in the style transfer output.}\label{fig: texture}
\end{figure*}

\subsection{Mapped Objects Loss}\label{ssec: mapped}
The mapped objects loss $\mathcal{L}^{M}_{dos}$ is computed between the output image, and the targeted style and the content images. $\mathcal{L}^{M}_{dos}$ is a combination of the deep photo-style loss \cite{luan2017deep} denoted by $\mathcal{L}_{dps}$ and contextual content loss denoted by $\mathcal{L}_{cl,C}$. We define $\mathcal{L}^{M}_{dos}$ in Eq.~\ref{eq: equal}.
\begin{equation}\label{eq: equal}
\mathcal{L}^{M}_{dos} = \alpha_1 {\mathcal{L}_{dps}} + \alpha_2 {\mathcal{L}_{cl,C}}
\end{equation}
Here, ${\mathcal{L}_{dps}}$ transfers features from the style image $S$ and the content image $C$ to the output image $O$. Whereas, ${\mathcal{L}_{cl,C}}$ transfers the content features from $C$ to $O$. 


The deep photo-style loss ${\mathcal{L}_{dps}}$ in Eq.~\ref{eq: equal} constrains image features of the style transfer output to be locally affine in colorspace to suppress distortions and yields photorealistic style transfer. The contextual content loss ${\mathcal{L}_{cl,C}}$ in Eq.~\ref{eq: equal} transfers the features by minimizing the dissimilarity between the contextually similar vectors sampled from the content representation $F_l[C]$ of $C$ and the content representation $F_l[O]$ of $O$. Here, $F_l[\cdot]$ denotes the vectorized feature map present at layer $l$ of features extractor $\phi(\cdot)$.  

The contextual content loss ${\mathcal{L}_{cl,C}}$ at a layer $l$ of $\phi$ is given as follows: ${\mathcal{L}^l_{cl,C}} =- \log CX(F_l[O], F_l[C]) $. Here, $CX(\cdot,\cdot)$ computes the contextual similarity \cite{mechrez2018contextual} between the content features of $C$ and $O$. Thus, it helps to preserve the semantics of the objects in output $O$. We have described deep photo-style loss and contextual loss in the supplementary material in detail.

The mapped objects loss $\mathcal{L}^{M}_{dos}$ defined in Eq.~\ref{eq: equal} is directly useful for STP-E as there are no unmapped objects (Fig.~\ref{fig: mappingPic}). Similarly, the image features transfer for mapped objects in STP-S and STP-C is done using $\mathcal{L}^{M}_{dos}$.

\begin{figure*}[!h] 
\begin{center}
\resizebox{\linewidth}{!}{%
\begin{subfigure}[b]{0.138\linewidth}\centering
    \includegraphics[width=\linewidth]{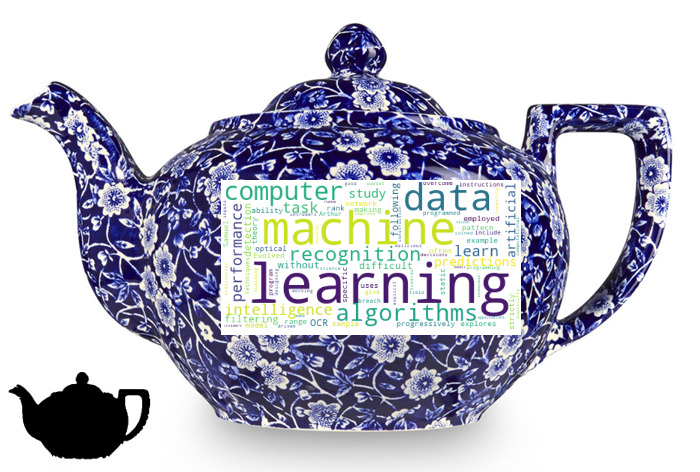}\\
\begin{minipage}{0.49\linewidth}\includegraphics[width=\linewidth]{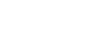}\end{minipage}%
\begin{minipage}{0.49\linewidth}\includegraphics[width=\linewidth]{new_images/Geomatry/in45/blank.jpg}\end{minipage}   \caption{\footnotesize  Style} 
    \end{subfigure} \hspace*{1pt}
\begin{subfigure}[b]{0.138\linewidth}\centering\includegraphics[width=\linewidth]{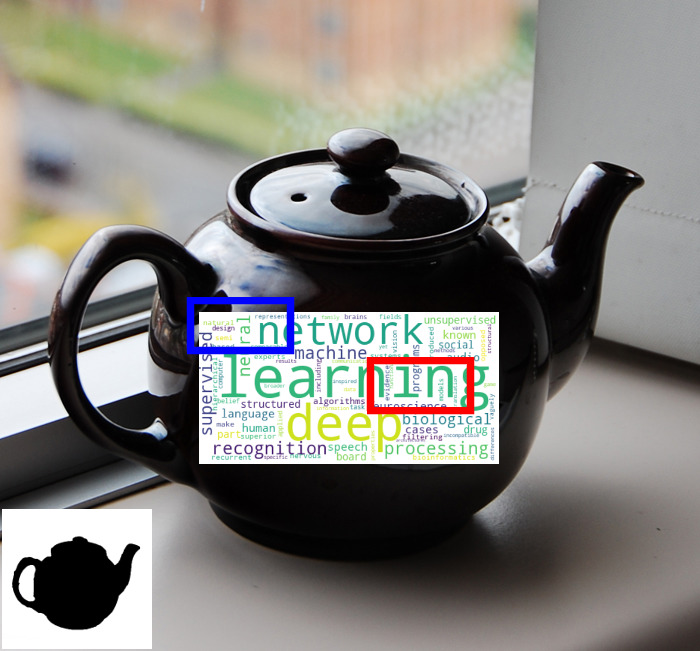} \\
\begin{minipage}{0.49\linewidth}\includegraphics[width=\linewidth, cfbox=blue 0.1pt 0.1pt]{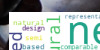}\end{minipage}%
\begin{minipage}{0.49\linewidth}\includegraphics[width=\linewidth, cfbox=red 0.1pt 0.1pt]{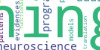}\end{minipage}%
\caption{\footnotesize  Content} 
    \end{subfigure} \hspace*{1pt}
\begin{subfigure}[b]{0.138\linewidth}\centering\includegraphics[width=\linewidth]{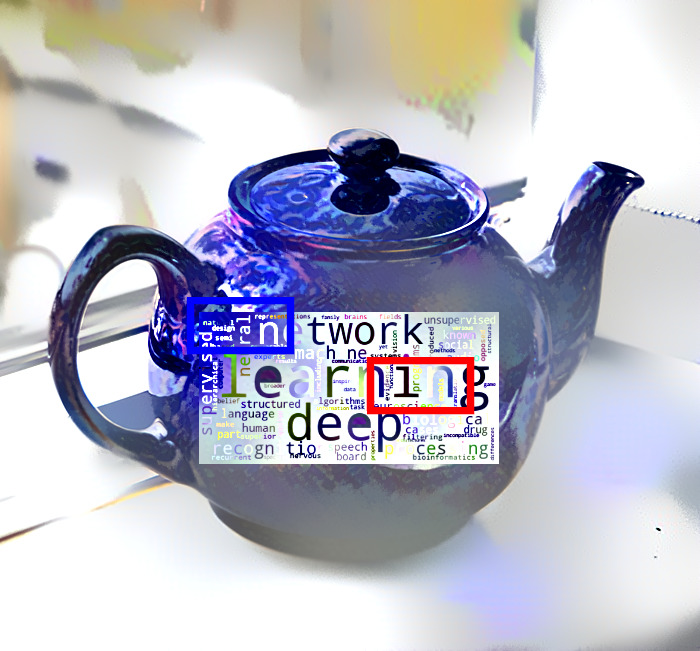} \\
\begin{minipage}{0.49\linewidth}\includegraphics[width=\linewidth, cfbox=blue 0.1pt 0.1pt]{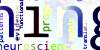}\end{minipage}%
\begin{minipage}{0.49\linewidth}\includegraphics[width=\linewidth, cfbox=red 0.1pt 0.1pt]{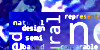}\end{minipage} \caption{\footnotesize  DPS  \cite{luan2017deep}} \end{subfigure}\hspace*{1pt}
\begin{subfigure}[b]{0.138\linewidth}\centering\includegraphics[width=\linewidth]{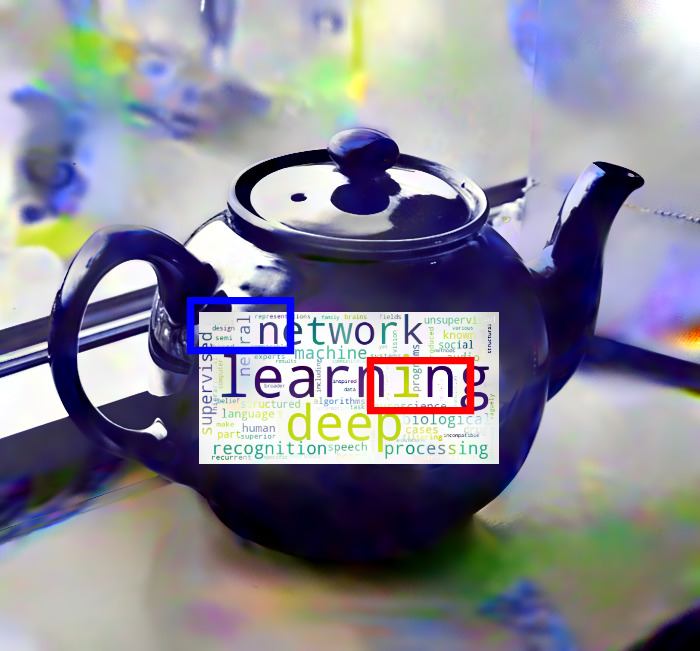} \\
\begin{minipage}{0.49\linewidth}\includegraphics[width=\linewidth, cfbox=blue 0.1pt 0.1pt]{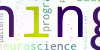}\end{minipage}%
\begin{minipage}{0.49\linewidth}\includegraphics[width=\linewidth, cfbox=red 0.1pt 0.1pt]{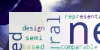}\end{minipage} \caption{\footnotesize  CL \cite{mechrez2018contextual}+$\mathcal{L}_m$\cite{luan2017deep}}\end{subfigure}\hspace*{1pt}
\begin{subfigure}[b]{0.138\linewidth}\centering\includegraphics[width=\linewidth]{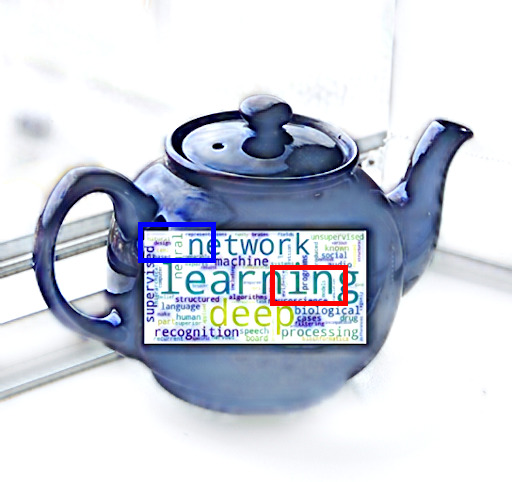} \\
\begin{minipage}{0.49\linewidth}\includegraphics[width=\linewidth, cfbox=blue 0.1pt 0.1pt]{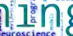}\end{minipage}%
\begin{minipage}{0.49\linewidth}\includegraphics[width=\linewidth, cfbox=red 0.1pt 0.1pt]{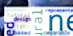}\end{minipage}\caption{\footnotesize  WCT2 \cite{yoo2019photorealistic}} \end{subfigure}\hspace*{1pt}
\begin{subfigure}[b]{0.138\linewidth}\centering\includegraphics[width=\linewidth]{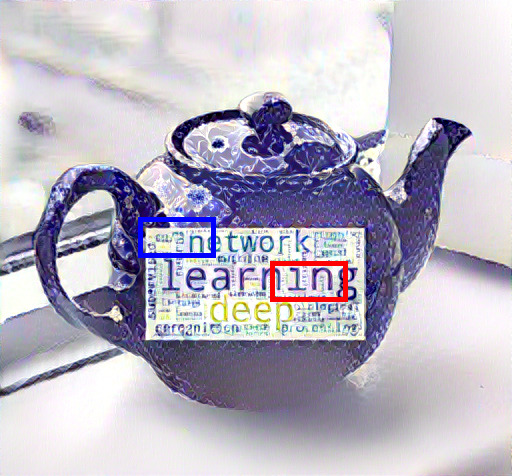} \\
\begin{minipage}{0.49\linewidth}\includegraphics[width=\linewidth, cfbox=blue 0.1pt 0.1pt]{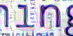}\end{minipage}%
\begin{minipage}{0.49\linewidth}\includegraphics[width=\linewidth, cfbox=red 0.1pt 0.1pt]{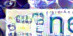}\end{minipage}\caption{\footnotesize  STROTSS  \cite{kolkin2019style}} \end{subfigure} \hspace*{1pt}%
\begin{subfigure}[b]{0.138\linewidth}\centering\includegraphics[width=\linewidth]{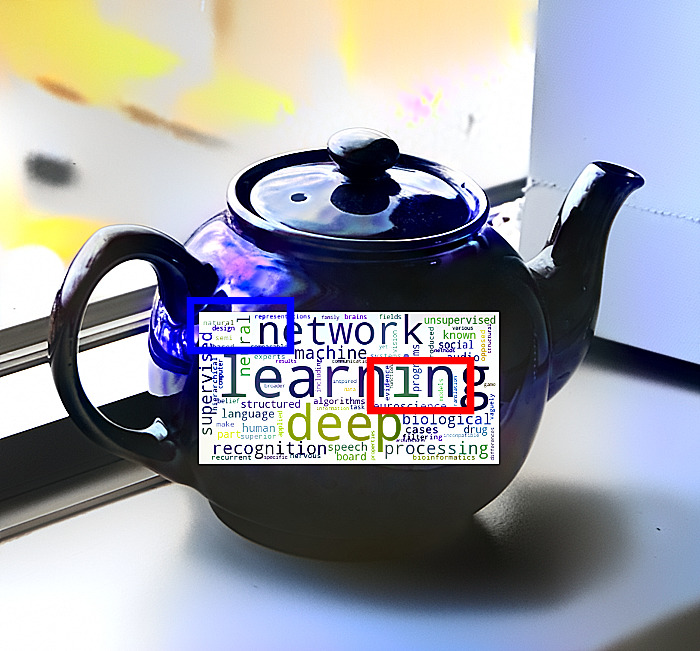} \\
\begin{minipage}{0.49\linewidth}\includegraphics[width=\linewidth, cfbox=blue 0.1pt 0.1pt]{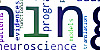}\end{minipage}%
\begin{minipage}{0.49\linewidth}\includegraphics[width=\linewidth, cfbox=red 0.1pt 0.1pt]{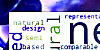}\end{minipage} \caption{\footnotesize  DeepObjStyle}\end{subfigure}} \end{center}
\vspace*{-0.3cm} \caption{\textbf{Preservation of Structure (STP-E).} The style image (a) and the content image (b) contains a word cloud. DPS \cite{luan2017deep} spills feature over the word cloud and suffers from the content mismatch. We integrated contextual loss \cite{mechrez2018contextual}  with photo-realism regularization $\mathcal{L}_m$ \cite{luan2017deep}. It shows a better word cloud region, but the features for other areas are not well distributed. WCT2 \cite{yoo2019photorealistic} and STROTSS \cite{kolkin2019style}  do not preserve the structure and the word cloud is less readable. DeepObjStyle (ours) minimizes content mismatch and preserves the geometry of the objects  {\color{blue} (the images are best viewed after zooming)}. } \label{fig: wordcloud}
\end{figure*}

\subsection{Unmapped Objects Loss}\label{ssec: unmapped}
Fig.~\ref{fig: mappingPic} shows the unmapped objects in two scenarios: STP-C and STP-S. The unmapped objects loss $\mathcal{L}^{U}_{dos}$ achieves style transfer for unmapped objects. $\mathcal{L}^{U}_{dos}$ is computed between the output image $O$ and the style image $S$. 

$\mathcal{L}^{U}_{dos}$ is a combination of the gram loss ${\mathcal{L}_{gl,S}}$, and contextual loss computed on style features, denoted by contextual style loss  ${\mathcal{L}_{cl,S}}$  as shown in Eq.~\ref{eq: differ}. 
\begin{equation}\label{eq: differ}
\mathcal{L}^{U}_{dos} = \beta_1 \underset{l\in L}{\sum} {\mathcal{L}_{gl,S}^l} + \beta_2 \underset{l\in L}{\sum} {\mathcal{L}_{cl,S}^l}
\end{equation}
Here, $\mathcal{L}_{cl,S}^l$ works by minimizing the difference in the contextually similar vectors in the feature representations of the style image and the output image at the layer $l$. $\mathcal{L}_{gl,S}^l$ uses the gram matrix of the feature maps to spread features.

For STP-C, the contextual style loss is computed between the style image and the unmapped objects present in the output image to achieve \emph{style diffusion}. For STP-S, the contextual style loss is computed between the unmapped objects in the style image $S$ and the output image $O$ to achieve \emph{style utilization}.

Let $\vartheta_c=\{ c_i\}_{i=1}^{m}$ be the set of segmentation channels in the content image $C$, where each segmentation channel represents an object of $C$. Similarly, let $\vartheta_s=\{s_i\}_{i=1}^{n}$ be the set of segmentation channels in the style image $S$. The style transfer for unmapped objects is described as follows. 


\textbf{Style diffusion (STP-C).} It is style transfer when the number of semantic objects in the content image $m$ is more than the number of semantic objects in the style image $n$ (\textit{i.e.}, $m>n$ in Fig.~\ref{fig: mappingPic}-b). There are a total of $(m+n)$ objects in $S$ and $C$, out of which $2n$ objects are mapped objects and $(m-n)$ unmapped objects.  We describe the strategy for style diffusion for $(m-n)$ unmapped objects in the output $O$ below. 

 First, we obtain the unmapped $(m-n)$ objects from $O$ by computing the Hadamard product between $O$ and the segmentation of unmapped $(m-n)$ objects $\vartheta_{c,n}$, where $\vartheta_{c,n} = {\sum}_{j=n}^{m} c_j$. Next, we provide the style features to the unmapped objects using the unmapped objects loss $\mathcal{L}^U_{dos}$ introduced earlier in Eq.~\ref{eq: differ}. For achieving this, the gram loss $\mathcal{L}_{gl,S}^l$ and the contextual style loss $\mathcal{L}_{cl,S}^l$ are defined in Eq.~\ref{eq: gramDiffusion} and Eq.~\ref{eq: contextualStyleDiffusion}. 
\begin{equation}\label{eq: gramDiffusion}
\mathcal{L}_{gl,S}^l = \frac{1}{2N^2_{l}} \underset{ij}{\sum} \big( G(F_{l}[O] \odot \vartheta_{c,n}) - G(F_{l} [S]) \big)^2_{ij} 
\end{equation}
Here, $N_l$ denotes the filters in layer $l$. Gram matrix $G[\cdot]$ is an inner product between vectorized style feature maps taken from $F_{l}[\cdot]$, where $F_l[\cdot]\in \mathbb{R}^{N_l \times D_l}$ and $D_l$ denotes the size of the vectorized feature map related to the filter. Thus, $G[\cdot] = F_l[\cdot]F_l[\cdot]^{T} \in \mathbb{R}^{N_l \times N_l} $. Gram loss in Eq.~\ref{eq: gramDiffusion} spreads style using the style representations $F_{l}[S]$ of the style image $S$ and the style representations $F_{l}[O]$ of the output image $O$. Eq.~\ref{eq: gramDiffusion} shows that we compute the feature correlations using the gram matrix of the style representation $G(F_{l}[S])$ and the representations of the unmapped objects of the output $G(F_{l}[O] \odot \vartheta_{c,n})$. The contextual style loss $\mathcal{L}_{cl,S}^l$ is given in Eq.~\ref{eq: contextualStyleDiffusion}.
\begin{equation}\label{eq: contextualStyleDiffusion}
\mathcal{L}_{cl,S}^l = - \log CX \big(F_{l}[O] \odot \vartheta_{c,n}, F_{l}[S] \big)  
\end{equation}
Here, $\mathcal{L}_{cl,S}^l$ works by minimizing the difference in the contextually similar vectors sampled from the style representation $F_{l}[S]$ of the style image $S$ and the style representation $F_{l}[O] \odot \vartheta_{c,n}$ of the unmapped objects in the output image $O$.

\textbf{Style utilization (STP-S).} It is style transfer when the number of semantic objects in the style image is more than that of the content image (\textit{i.e.}, $n>m$ in Fig.~\ref{fig: mappingPic}-c). The challenge is to achieve style utilization of the $(n-m)$ unmapped objects in the style image using the loss $\mathcal{L}^U_{dos}$ introduced earlier in Eq.~\ref{eq: differ}. 

First, we obtain the $(n-m)$ unmapped objects in the style image using their segmentation mask $\vartheta_{s,m}$, where $\vartheta_{s,m} = {\sum}_{j=m}^{n} s_j$.  Next, we provide the style features to them using the gram loss $\mathcal{L}_{gl,S}^l$ and the contextual loss $\mathcal{L}_{cl,S}^l$ defined in Eq.~\ref{eq: gramUtilization} and Eq.~\ref{eq: contextualStyleUtilization}. 

\begin{equation}\label{eq: gramUtilization}
\mathcal{L}_{gl,S}^l = \frac{1}{2N^2_{l}} \underset{ij}{\sum} \big( G(F_{l}[O]) - G(F_{l}[S] \odot \vartheta_{s,m}) \big)^2_{ij} 
\end{equation}
\begin{equation}\label{eq: contextualStyleUtilization} 
\mathcal{L}_{cl,S}^l = - \log CX \big(F_{l}[O], F_{l}[S] \odot \vartheta_{s,m} \big)  
\end{equation}
Eq.~\ref{eq: gramUtilization} and Eq.~\ref{eq: contextualStyleUtilization} show that we use the features from the unmapped objects in style utilization similar to style diffusion (Eq.~\ref{eq: gramDiffusion} and Eq.~\ref{eq: contextualStyleDiffusion}). In the supplementary material, we provide more technical details for the unmapped objects.

\section{Experimental Results} \label{sec: experiments}
Fig.~\ref{fig: photo} and Fig.~\ref{fig: texture} show the photo-realistic style transfer when the style and the content images have different categories of the objects. We experiment with the content images that are photo-realistic and the task is to preserve the object structure properties. In Fig.~\ref{fig: photo}, the style image is photo-realistic. In Fig.~\ref{fig: texture}, the style image is not photo-realistic. The main intuition is that images with less deformation are more visually appealing. Neural style \cite{gatys2016image} deforms the geometry of the image. DPS \cite{luan2017deep} preserves object boundaries, but the style features are not distributed well. WCT2 \cite{yoo2019photorealistic} does not preserve fine image feature details in the style transfer output. STROTSS \cite{kolkin2019style} does not distribute image features well. DeepObjStyle is able to preserve the photo-realistic properties of the content image in both the scenarios shown in Fig.~\ref{fig: photo} and Fig.~\ref{fig: texture}. 

Fig.~\ref{fig: mismatch} shows the style transfer when the style and the content images have an extreme mismatch of image features. DPS \cite{luan2017deep}, WCT2 \cite{yoo2019photorealistic}, and STROTSS \cite{kolkin2019style} output images with higher perceptual error and lower image quality score. DPS \cite{luan2017deep} does not preserve the semantics of the objects. WCT2 \cite{yoo2019photorealistic} trains a decoder on sample images. Therefore, the higher perceptual error might be because of the bias toward the sample images and lacking the generalization to the new images. STROTSS \cite{kolkin2019style} transport style features onto the content image with minimum distortion to the geometry of the objects, but in the challenging scenario of content mismatch, the structure preservation reduces. DeepObjStyle outperforms other methods and preserves the semantics of the objects in the output. 

Fig.~\ref{fig: wordcloud} shows the style transfer when the style and content images contain a word cloud. The challenge here is to supervise style and content features while maintaining the readability of the text. DPS \cite{luan2017deep} spills-over unrelated features on the word-cloud. To investigate the photo-realistic style supervision with contextual loss CL \cite{mechrez2018contextual}, we integrate the photo-realism regularization module $\mathcal{L}_m$ \cite{luan2017deep} with CL \cite{mechrez2018contextual}. The photo-realism mostly suppresses distortions and preserves the structure of the objects in the output. $\mathcal{L}_m$ \cite{luan2017deep} with CL \cite{mechrez2018contextual} does not distribute image features well. It might be because contextually similar features between the source and the target images were not well used in the output. WCT2 \cite{yoo2019photorealistic} and STROTSS \cite{kolkin2019style} does not preserve the local level image features details and reduce the text readability. DeepObjStyle provides a better distribution of features even when the segmentation mask does not provide the position of the word cloud\footnote{We illustrate the extended version of Fig.~\ref{fig: wordcloud} in the supplementary material.}. 

\begin{figure*}[!h]  \begin{center}  \resizebox{0.88\linewidth}{!}{%
\begin{subfigure}{0.145\linewidth}\begin{center}\includegraphics[width=\textwidth]{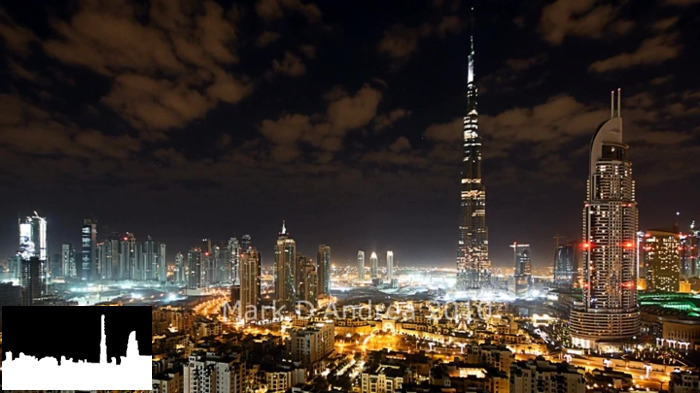} \\
\begin{subfigure}{0.5\textwidth}\includegraphics[width=\textwidth]{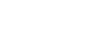}\end{subfigure}%
\begin{subfigure}{0.5\textwidth}\includegraphics[width=\textwidth]{new_images/StyleDiffusion/blank.jpg}\end{subfigure} \caption{Style $S$}\end{center} \end{subfigure}
\begin{subfigure}{0.145\linewidth}\begin{center}\includegraphics[width=\textwidth]{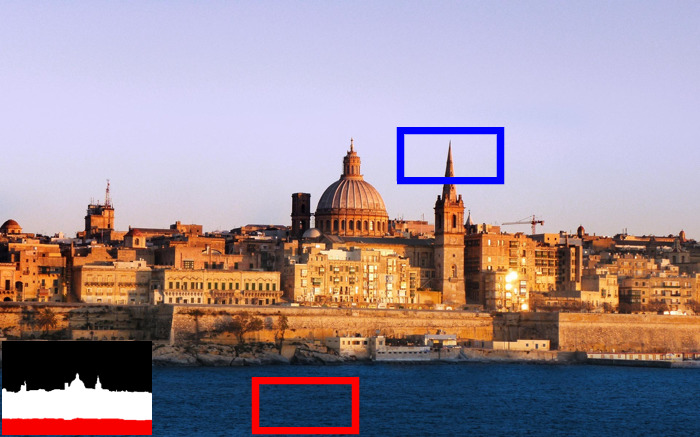} \\
\begin{subfigure}{0.5\textwidth}\includegraphics[width=\textwidth, cfbox=red 0.1pt 0.1pt]{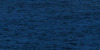}\end{subfigure}%
\begin{subfigure}{0.5\textwidth}\includegraphics[width=\textwidth, cfbox=blue 0.1pt 0.1pt]{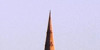}\end{subfigure} \caption{Content $C$}\end{center} \end{subfigure}
\begin{subfigure}{0.145\linewidth}\begin{center}\includegraphics[width=\textwidth]{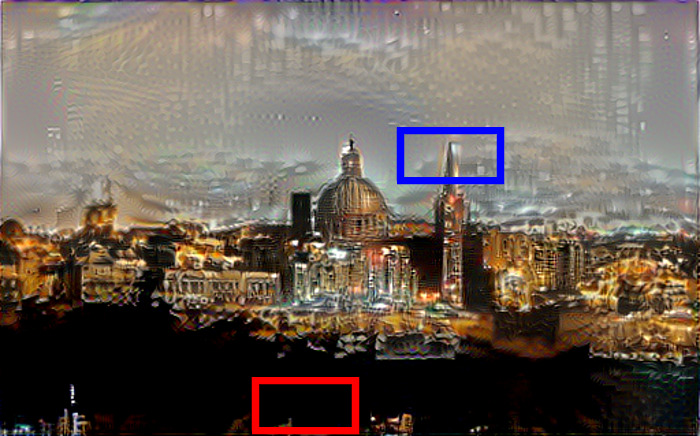} \\
\begin{subfigure}{0.5\textwidth}\includegraphics[width=\textwidth, cfbox=red 0.1pt 0.1pt]{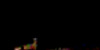} \end{subfigure}%
\begin{subfigure}{0.5\textwidth}\includegraphics[width=\textwidth, cfbox=blue 0.1pt 0.1pt]{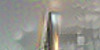}\end{subfigure}\caption{Neural style\cite{gatys2016image} }\end{center} \end{subfigure}
\begin{subfigure}{0.145\linewidth}\begin{center}\includegraphics[width=\textwidth]{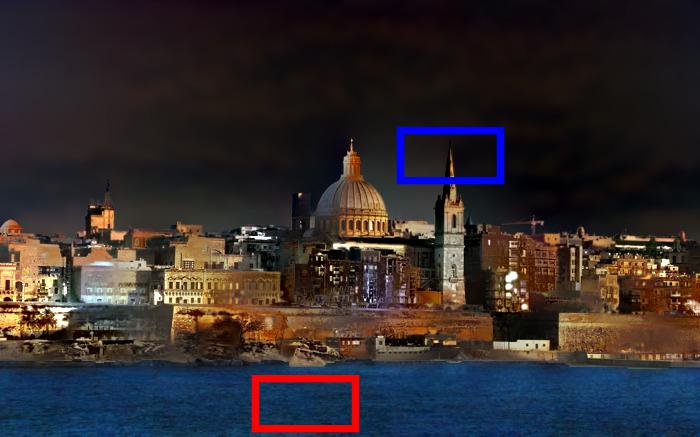} \\
\begin{subfigure}{0.5\textwidth}\includegraphics[width=\textwidth, cfbox=red 0.1pt 0.1pt]{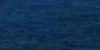}\end{subfigure}%
\begin{subfigure}{0.5\textwidth}\includegraphics[width=\textwidth, cfbox=blue 0.1pt 0.1pt]{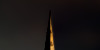}\end{subfigure}\caption{DPS \cite{luan2017deep} }\end{center} \end{subfigure}
\begin{subfigure}{0.145\linewidth}\begin{center}\includegraphics[width=\textwidth]{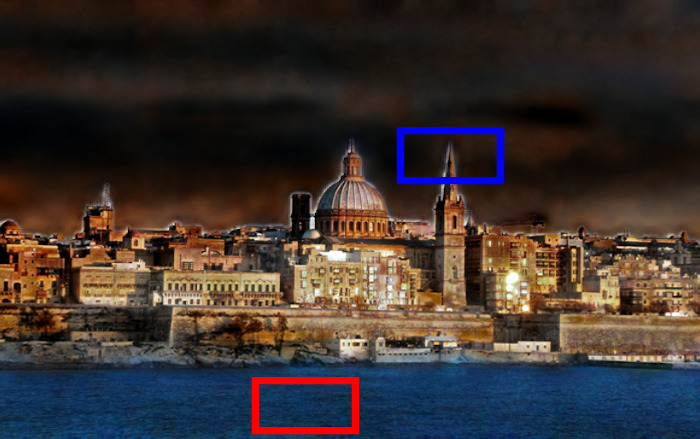} \\
\begin{subfigure}{0.5\textwidth}\includegraphics[width=\textwidth, cfbox=red 0.1pt 0.1pt]{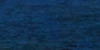}\end{subfigure}%
\begin{subfigure}{0.5\textwidth}\includegraphics[width=\textwidth, cfbox=blue 0.1pt 0.1pt]{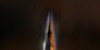}\end{subfigure}\caption{WCT2 \cite{yoo2019photorealistic} }\end{center} \end{subfigure}
\begin{subfigure}{0.145\linewidth}\begin{center}\includegraphics[width=\textwidth]{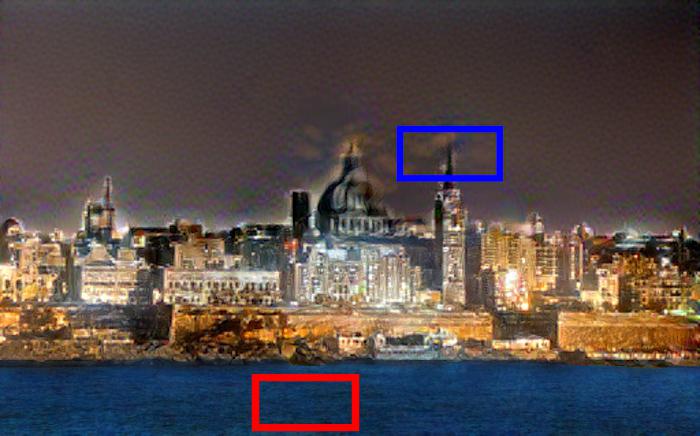} \\
\begin{subfigure}{0.49\textwidth}\includegraphics[width=\linewidth, cfbox=blue 0.1pt 0.1pt]{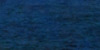}\end{subfigure}%
\begin{subfigure}{0.49\textwidth}\includegraphics[width=\linewidth, cfbox=red 0.1pt 0.1pt]{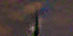}\end{subfigure} \caption{STROTSS  \cite{kolkin2019style}}\end{center} \end{subfigure}  \hspace*{0.1pt}%
\begin{subfigure}{0.145\linewidth}\begin{center}\includegraphics[width=\textwidth]{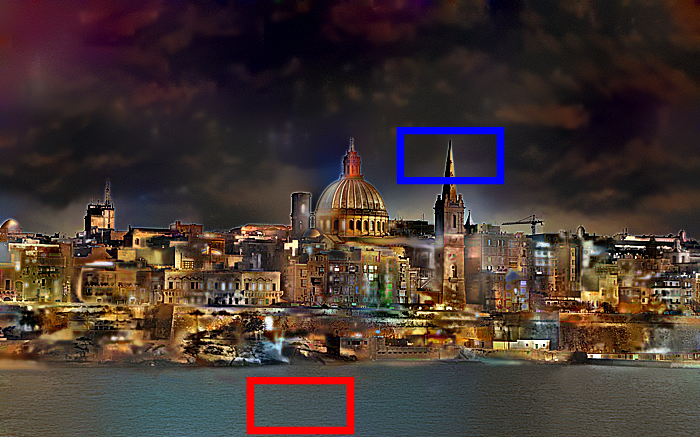} \\
\begin{subfigure}{0.5\textwidth}\includegraphics[width=\textwidth, cfbox=red 0.1pt 0.1pt]{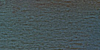}\end{subfigure}%
\begin{subfigure}{0.5\textwidth}\includegraphics[width=\textwidth, cfbox=blue 0.1pt 0.1pt]{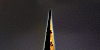}\end{subfigure} \caption{DeepObjStyle} 
\end{center} \end{subfigure} } \end{center}\vspace*{-0.3cm} \caption{\textbf{Style Diffusion (STP-C).} The content image has more semantic objects than that of style image. The semantic objects of the style and content images are shown by the segmentation mask at the bottom left corner. The sky and the building are the mapped objects. The unmapped content object is the lake, which is shown by red color in the segmentation mask in (b). Neural style \cite{gatys2016image} distorts the geometry of the objects. DPS \cite{luan2017deep} does not provide the style features to the unmapped lake object, thus style diffusion not achieved.  WCT2 \cite{yoo2019photorealistic} and STROTSS \cite{kolkin2019style} do not distribute image features well for the sky. DeepObjStyle (ours) achieves style diffusion while preserving the geometry of the objects {\color{blue} (the images are best viewed after zooming)}.}\label{fig: diffusion} 
\end{figure*}

\begin{figure*}[!h] \begin{center} 
\resizebox{0.88\linewidth}{!}{%
\begin{subfigure}{0.135\linewidth}\begin{center}\includegraphics[width=\textwidth]{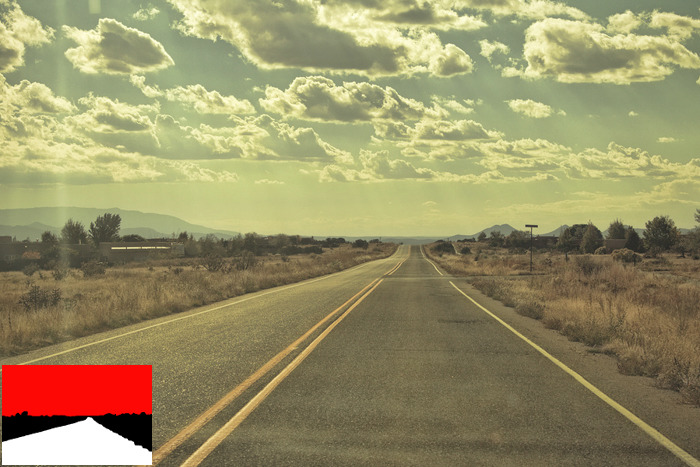} \\
\begin{subfigure}{0.5\textwidth}\includegraphics[width=\textwidth]{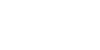}\end{subfigure}%
\begin{subfigure}{0.5\textwidth}\includegraphics[width=\textwidth]{new_images/StyleUtilization/blank.jpg}\end{subfigure} \caption{Style $S$}\end{center} \end{subfigure}
\begin{subfigure}{0.135\linewidth}\begin{center}\includegraphics[width=\textwidth]{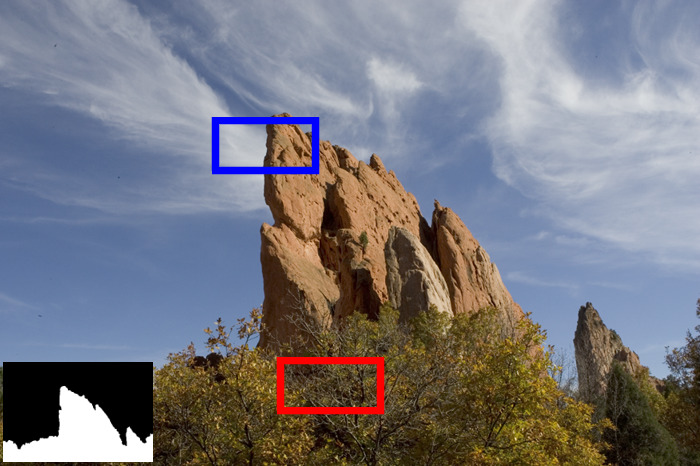} \\
\begin{subfigure}{0.5\textwidth}\includegraphics[width=\textwidth, cfbox=red 0.1pt 0.1pt]{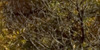}\end{subfigure}%
\begin{subfigure}{0.5\textwidth}\includegraphics[width=\textwidth, cfbox=blue 0.1pt 0.1pt]{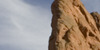}\end{subfigure}\caption{Content $C$}\end{center} \end{subfigure}
\begin{subfigure}{0.135\linewidth}\begin{center}\includegraphics[width=\textwidth]{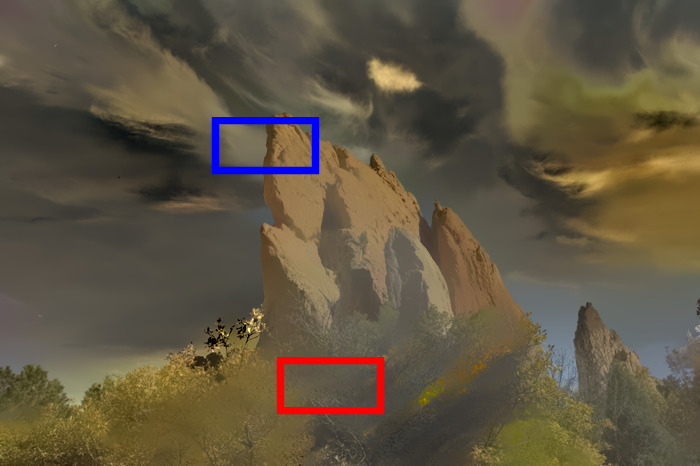} \\
\begin{subfigure}{0.5\textwidth}\includegraphics[width=\textwidth, cfbox=red 0.1pt 0.1pt]{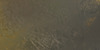}\end{subfigure}%
\begin{subfigure}{0.5\textwidth}\includegraphics[width=\textwidth, cfbox=blue 0.1pt 0.1pt]{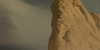}\end{subfigure}\caption{DPS \cite{luan2017deep}}\end{center} \end{subfigure}
\begin{subfigure}{0.135\linewidth}\begin{center}\includegraphics[width=\textwidth]{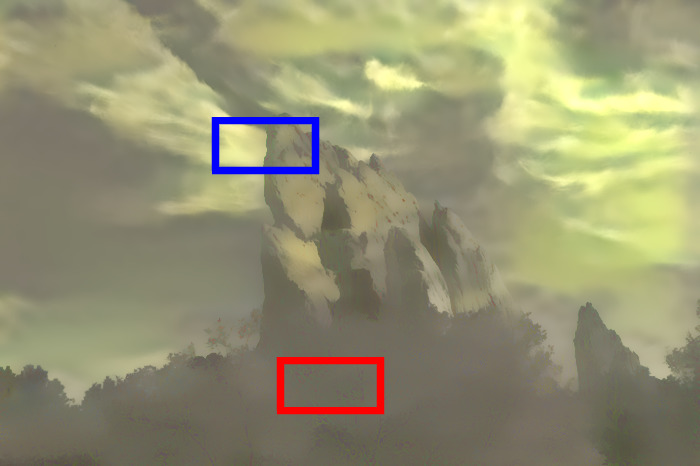} \\
\begin{subfigure}{0.5\textwidth}\includegraphics[width=\textwidth, cfbox=red 0.1pt 0.1pt]{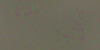}\end{subfigure}%
\begin{subfigure}{0.5\textwidth}\includegraphics[width=\textwidth, cfbox=blue 0.1pt 0.1pt]{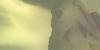}\end{subfigure} \caption{DPS$^+$}\end{center} \end{subfigure}
\begin{subfigure}{0.135\linewidth}\begin{center}\includegraphics[width=\textwidth]{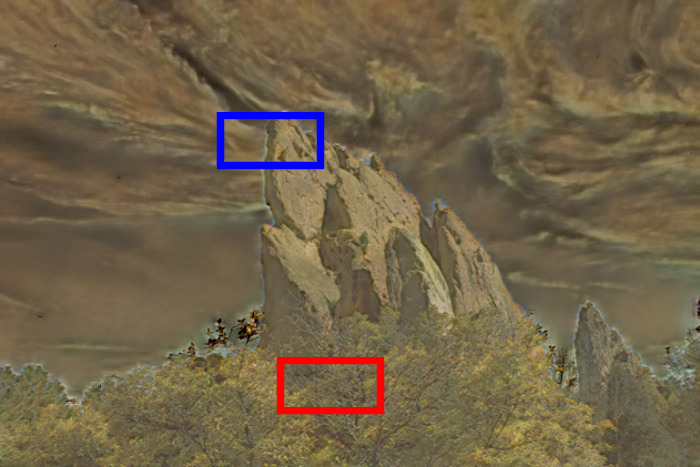} \\
\begin{subfigure}{0.5\textwidth}\includegraphics[width=\textwidth, cfbox=red 0.1pt 0.1pt]{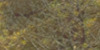}\end{subfigure}%
\begin{subfigure}{0.5\textwidth}\includegraphics[width=\textwidth, cfbox=blue 0.1pt 0.1pt]{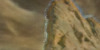}\end{subfigure} \caption{WCT2 \cite{yoo2019photorealistic} }\end{center} \end{subfigure}
\begin{subfigure}{0.135\linewidth}\begin{center}\includegraphics[width=\textwidth]{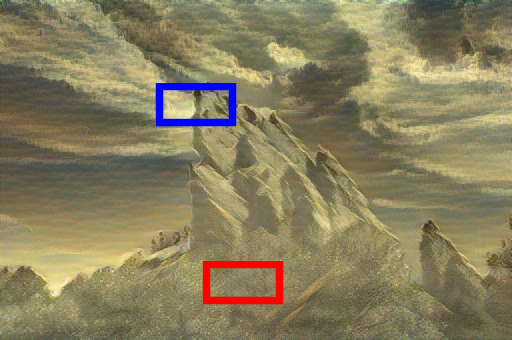} \\
\begin{subfigure}{0.49\textwidth}\includegraphics[width=\linewidth, cfbox=blue 0.1pt 0.1pt]{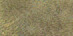}\end{subfigure}%
\begin{subfigure}{0.49\textwidth}\includegraphics[width=\linewidth, cfbox=red 0.1pt 0.1pt]{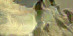}\end{subfigure}  \caption{STROTSS  \cite{kolkin2019style}}\end{center} \end{subfigure} 
\begin{subfigure}{0.135\linewidth}\begin{center}\includegraphics[width=\textwidth]{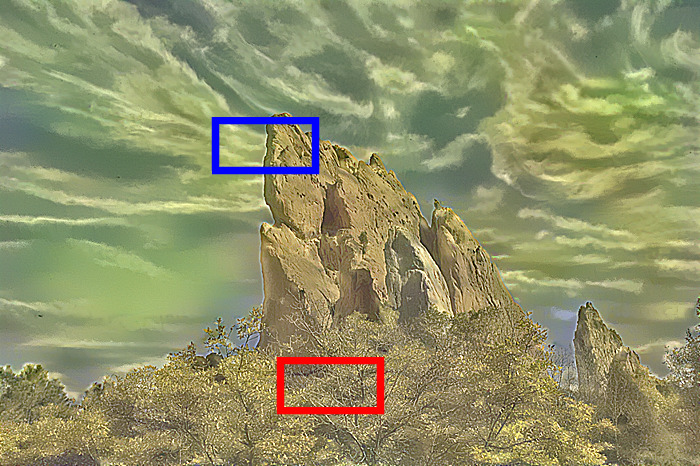} \\
\begin{subfigure}{0.5\textwidth}\includegraphics[width=\textwidth, cfbox=red 0.1pt 0.1pt]{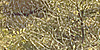}\end{subfigure}%
\begin{subfigure}{0.5\textwidth}\includegraphics[width=\textwidth, cfbox=blue 0.1pt 0.1pt]{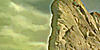}\end{subfigure}\caption{DeepObjStyle}\end{center} \end{subfigure}} \end{center}
\vspace*{-0.3cm}  \caption{\textbf{Style Utilization (STP-S).} The style image $S$ has more semantic objects than that of the content image $C$. We map the grass in $S$ with the sky in $C$ to show the effects of content mismatch. DPS \cite{luan2017deep} does not achieve style utilization as the style features of the sky are not present in the output image. We integrated unmapped object loss in DPS and called it DPS$^+$, for a fair comparison. DPS$^+$  achieves style utilization as we could see the color of the sky from $S$, but object boundaries are not visible. The image features are not much clear for WCT2 \cite{yoo2019photorealistic} and STROTSS \cite{kolkin2019style}. DeepObjStyle achieves style utilization as the color of the sky is also taken from the sky of $S$. DeepObjStyle learns the object context and preserves the geometry of the objects in the output {\color{blue} (see the cropped images)}.} \label{fig: utilization}
\end{figure*}

Fig.~\ref{fig: diffusion} shows style transfer when the content image has more objects than that of the style image. It is worth noting that merging lake object with sky or building would result in a high content mismatch. Neural style \cite{gatys2016image} spreads the style features disregarding the object boundaries. DPS \cite{luan2017deep} does not supervise features of the sky and the lake object well. WCT2  \cite{yoo2019photorealistic} does not fully utilize style features well, but it preserved better image features than STROTSS \cite{kolkin2019style}. DeepObjStyle spreads the style features to all the objects in the output image while preserving the structure.


Fig.~\ref{fig: utilization} shows the influence of content mismatch in style transfer when style image has more objects than that of content image. We purposefully mapped the sky of content image with trees of style image to make a challenging scenario for style transfer. DPS \cite{luan2017deep} does not preserve the content features well. To get a fair comparison, we improvise DPS with the unmapped objects loss in DPS$^+$ method (we describe DPS$^+$ in the supplementary material). It utilizes features from all the semantic objects of the style and the content images. However, the preservation of the object structure is not achieved. WCT2 \cite{yoo2019photorealistic} and STROTSS \cite{kolkin2019style} performs better but local image features details are not preserved. It can be observed that DeepObjStyle achieves style utilization while preserving the structure of the object. 




\section{Quantitative Comparision} \label{sec: quantitative}
\begin{figure}[!t]\centering
\includegraphics[width=0.9\linewidth]{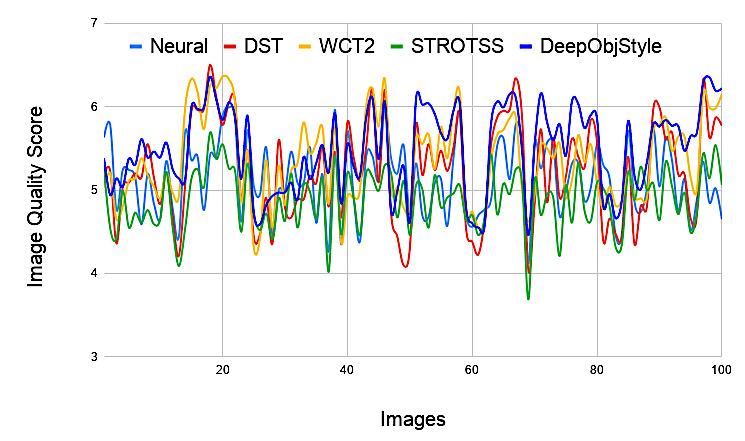}\vspace*{-0.15cm}
\caption{\textbf{Image Quality Comparision.} The figure shows the comparison of image quality of style transfer output using NIMA \cite{idealods2018imagequalityassessment}. DeepObjStyle achieves a better quality score as compared to other methods.}\label{fig: IQA_nima}
\end{figure}

Fig.~\ref{fig: IQA_nima} shows the no-reference quality assessment using NIMA \cite{idealods2018imagequalityassessment} for 100 style transfer instances. NIMA \cite{idealods2018imagequalityassessment} predicts the image quality score. The average image quality scores are as follows. Neural Style \cite{gatys2016image}: $5.13$, DPS \cite{luan2017deep}: $5.23$, WCT2 \cite{yoo2019photorealistic}: $5.35$, STROTSS \cite{kolkin2019style}: $4.88$, and DeepObjStyle: $5.49$. DeepObjStyle outperforms other methods by max image quality score.

\begin{figure}[!t]\centering
\includegraphics[width=0.9\linewidth]{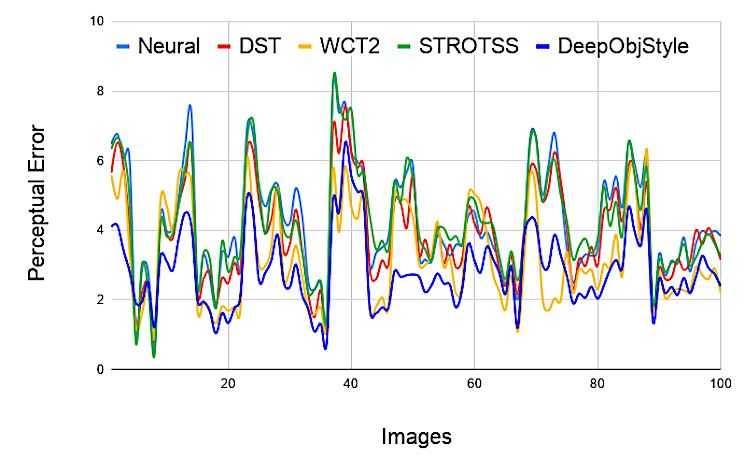}\vspace*{-0.15cm}
\caption{\textbf{Perceptual Error Comparison.} The figure shows perceptual error comparison with the reference of input content image using Pieapp \cite{prashnani2018pieapp}. DeepObjStyle shows minimum distortion in the output as compared to other methods. Therefore, output images of DeepObjStyle are more visually appealing.}\label{fig: IQA_pieapp}
\end{figure}

Fig.~\ref{fig: IQA_pieapp} shows the reference-based quality assessment using perceptual error scores Pieapp \cite{prashnani2018pieapp} to investigate the distortion of the content features. The average Pieapp \cite{prashnani2018pieapp} scores are as follows. Neural Style \cite{gatys2016image}: $4.26$, DPS \cite{luan2017deep}: $3.92$, WCT2 \cite{yoo2019photorealistic}: $3.22$, STROTSS \cite{kolkin2019style}: $4.21$, and DeepObjStyle: $2.83$. DeepObjStyle outperforms other methods by minimum perceptual error.

We have also conducted a user study to validate the style transfer results. We took 18 style transfer instances and displayed them in random order. Each subject is asked to vote for the better-looking image. We collected feedback from 60 human experts with a total of 1080 votes. The number of votes for the methods is as follows. Neural Style \cite{gatys2016image}: 72, DPS \cite{luan2017deep}: 78, WCT2 \cite{yoo2019photorealistic}: 174, STROTSS \cite{kolkin2019style}: 76, and DeepObjStyle: 680 (highest votes). 

The user study, perceptual error score \cite{prashnani2018pieapp}, and image quality score \cite{idealods2018imagequalityassessment} confirms the observation that DeepObjStyle transfer image features with better structure preservation and lesser distortions are visually more appealing\footnote{We used the implementation of Neural style provided in \cite{Smith2016}, Tensorflow implementation of DPS given in \cite{YangPhotoStyle2017}, contextual loss implementation in \cite{roimehrez2018}, STROTSS implementation in \cite{nkolkin13STROTSS}, and WCT2 implementation in \cite{clovaaiWCT2}. We have provided more visual comparisons and implementation details of our method in the supplementary material.}. 

\begin{figure}[!t]\centering
\begin{subfigure}[b]{0.3\linewidth} \includegraphics[width=\linewidth]{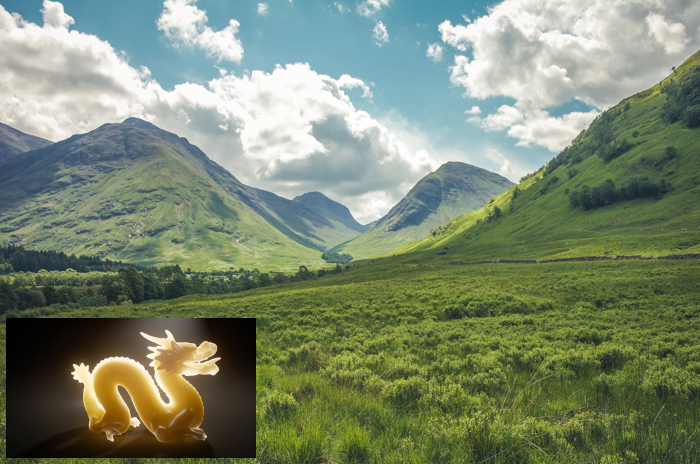} \caption{\small  Content \& Style} \end{subfigure}\hspace*{1pt} 
\begin{subfigure}[b]{0.3\linewidth} \includegraphics[width=\linewidth]{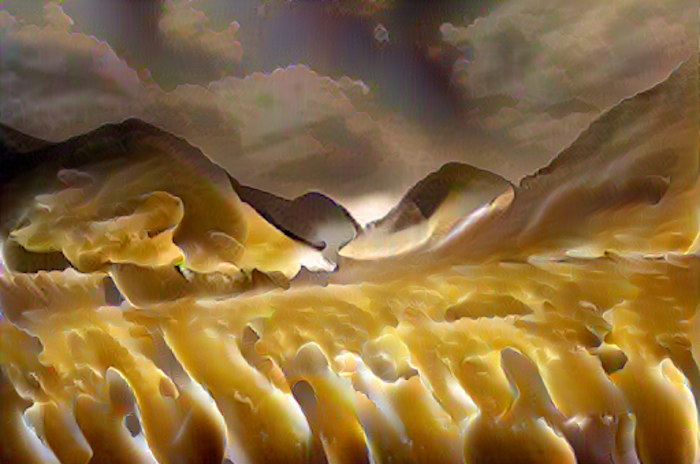}\caption{\small  STROTSS \cite{kolkin2019style}}\end{subfigure}\hspace*{1pt} 
\begin{subfigure}[b]{0.3\linewidth} \includegraphics[width=\linewidth]{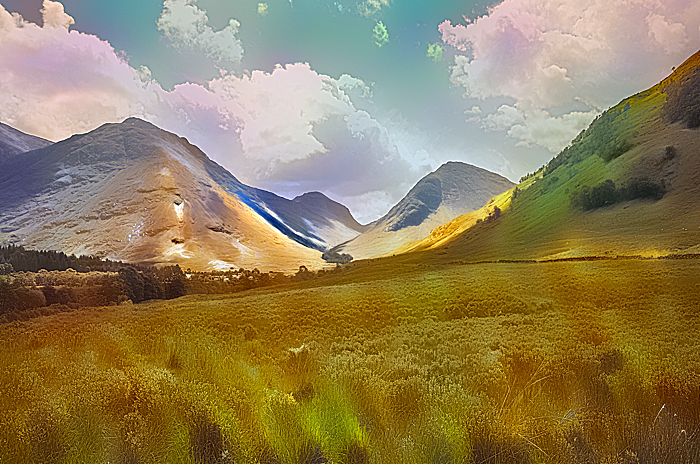}\caption{\small  DeepObjStyle}\end{subfigure}\vspace*{-0.15cm}
\caption{\textbf{Limitation-1.} DeepObjStyle preserves a better object context, but the distribution of features might be improved further. }\label{fig: failEx1}
\end{figure}%
\begin{figure}[!t]\centering
\begin{subfigure}[b]{0.3\linewidth} \includegraphics[width=\linewidth]{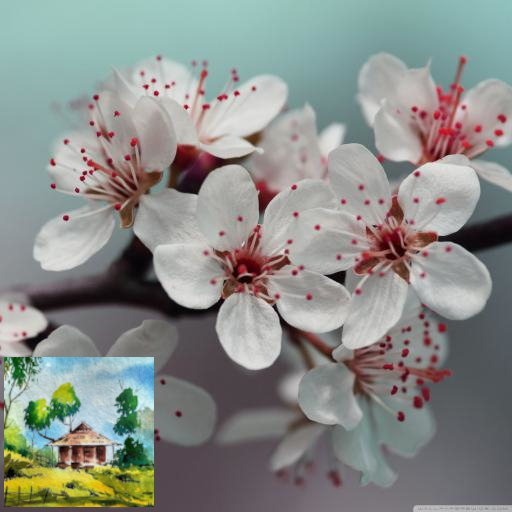} \caption{\small  Content \& Style} \end{subfigure}
\begin{subfigure}[b]{0.3\linewidth} \includegraphics[width=\linewidth]{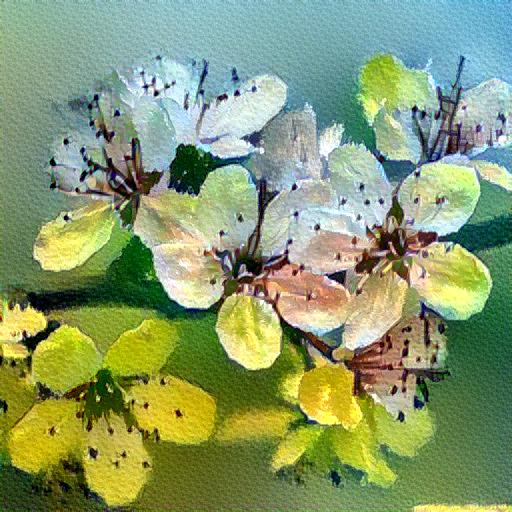} \caption{\small  STROTSS \cite{kolkin2019style}} \end{subfigure}
\begin{subfigure}[b]{0.3\linewidth} \includegraphics[width=\linewidth]{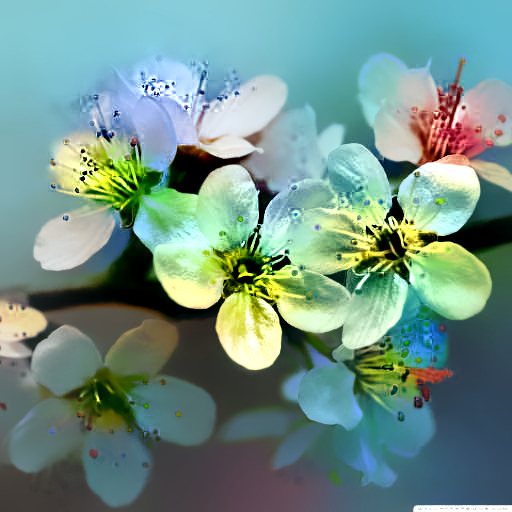} \caption{\small  DeepObjStyle} \end{subfigure}\vspace*{-0.15cm}
\caption{\textbf{Limitation-2.} DeepObjStyle minimizes the distortion. Therefore, it misses fine texture information. STROTSS \cite{kolkin2019style} does not preserve photo-realism. A similar effect is shown in the second row of Fig.~\ref{fig: mismatch}. }\label{fig: failEx2}
\end{figure}

\section{Limitationss}
Fig.~\ref{fig: failEx1} and Fig.~\ref{fig: failEx2} show the limitations of DeepObjStyle. Fig.~\ref{fig: failEx1} shows the style transfer when style and content images have an extreme mismatch of features. Fig.~\ref{fig: failEx2} shows the style transfer when the style image is an artistic image with painting style deformations. The limitations are due to the DeepObjStyle strategy of contextually similar feature comparison and minimizing deformations. We observed that mixing uncorrelated image features in the style transfer output results in a high perceptual error and low-quality output in general (Sec.~\ref{sec: quantitative}).


\section{Conclusion} \label{sec: conclusion} 
DeepObjStyle achieves good image features supervision in many challenging scenarios, such as content mismatch and style transfer of images containing a word cloud. DeepObjStyle achieves style diffusion and style utilization while minimizing content mismatch. DeepObjStyle preserves the semantics of the objects and the object structure. The extensive experiments and quality assessment shows that DeepObjStlye outperforms relevant style transfer methods. We believe that perceptual quality in the extreme content mismatch scenario could be enhanced further. We propose designing a method that performs style transfer in the extreme content mismatch scenario as future work.

{\small
\bibliographystyle{ieee_fullname}
\bibliography{egbib}
}

\end{document}